\documentclass[letterpaper, 10 pt, conference]{ieeeconf}  

\IEEEoverridecommandlockouts                              

\usepackage{graphicx}
\usepackage{makecell}   
\usepackage{pifont}     
\usepackage[table]{xcolor} 
\usepackage{multirow}
\usepackage{wrapfig}
\usepackage{amsmath}
\usepackage{arydshln}
\usepackage{subcaption}
\definecolor{dropcolor}{RGB}{234, 107, 21}
\usepackage{booktabs}  
\usepackage{amsfonts}
\usepackage{hyperref}

\title{\LARGE \bf
RGBD20K: A Large-Scale Benchmark for RGB-D
Semantic Segmentation
}

\author{Shaohua Dong, Zexuan Meng, Haiyan Sun, Bing Fan, \\ 
Cuicui Zhang, Dylan Joseph, Kewei Sha, Yunhe Feng and Heng Fan
\thanks{The authors are with the University of North Texas. Denton, TX 76207, USA.
{shaohuadong@my.unt.edu, heng.fan@unt.edu}
}}

\begin{document}

\maketitle
\thispagestyle{empty}
\pagestyle{empty}

\begin{abstract}

In this paper, we propose \textbf{RGBD20K}, a novel dataset for facilitating the development of more robust and general RGB-D semantic segmentation by encompassing abundant categories and high-quality annotations. RGBD20K possesses several attractive properties: (1) \textbf{\emph{Expanded Semantic Space}.} In particular, it covers 160 fine-grained categories, largely surpassing the category diversity of existing popular RGB-D benchmarks (e.g., NYUv2 with 40 classes and SUN RGB-D with 37 classes). With such enriched semantic coverage, we expect to promote the learning of more generalizable segmentation models. (2) \textbf{\emph{Larger Scale}.} Compared with current benchmarks, RGBD20K offers 20,000 RGB-D image pairs, providing a substantially larger training resource that benefits the development of more powerful deep models. (3) \textbf{\emph{High-Fidelity Annotation}.} We perform rigorous re-evaluation and correction of existing labels to resolve long-standing annotation noise, resulting in a clean and reliable ground-truth foundation. Furthermore, we propose a novel \textbf{\emph{score-purified fusion}} (SPF) method, which achieves state-of-the-art performance across all evaluated benchmarks, demonstrating the effectiveness of our approach in leveraging high-quality multimodal information for RGB-D semantic segmentation. The dataset is here: \href{https://github.com/ShaohuaDong2021/RGBD20K/}{RGBD20K}.

\end{abstract}

\section{INTRODUCTION}

Visual perception \cite{xie2021segformer, peng2024vasttrack, dong2024loretrack, li2025dmtrack} is a fundamental problem in computer vision, with semantic segmentation serving as a core task for achieving dense and structured scene understanding. It has been widely applied in robotics, autonomous systems, and intelligent perception, where pixel-level recognition of complex scenes is essential. Despite significant progress in deep learning-based RGB-D semantic segmentation, current methods \cite{dong2024efficient, yin2023dformer, yin2025dformerv2} are still far from achieving robust and generalizable performance in real-world environments. A key limiting factor lies not only in model design, but more fundamentally in severe limitations of existing RGB-D datasets, as described in the following.

\begin{figure*}[!t]
    \centering
    \setlength{\tabcolsep}{3pt}
    \begin{subfigure}[b]{0.31\textwidth}
        \centering
        \includegraphics[width=\textwidth]{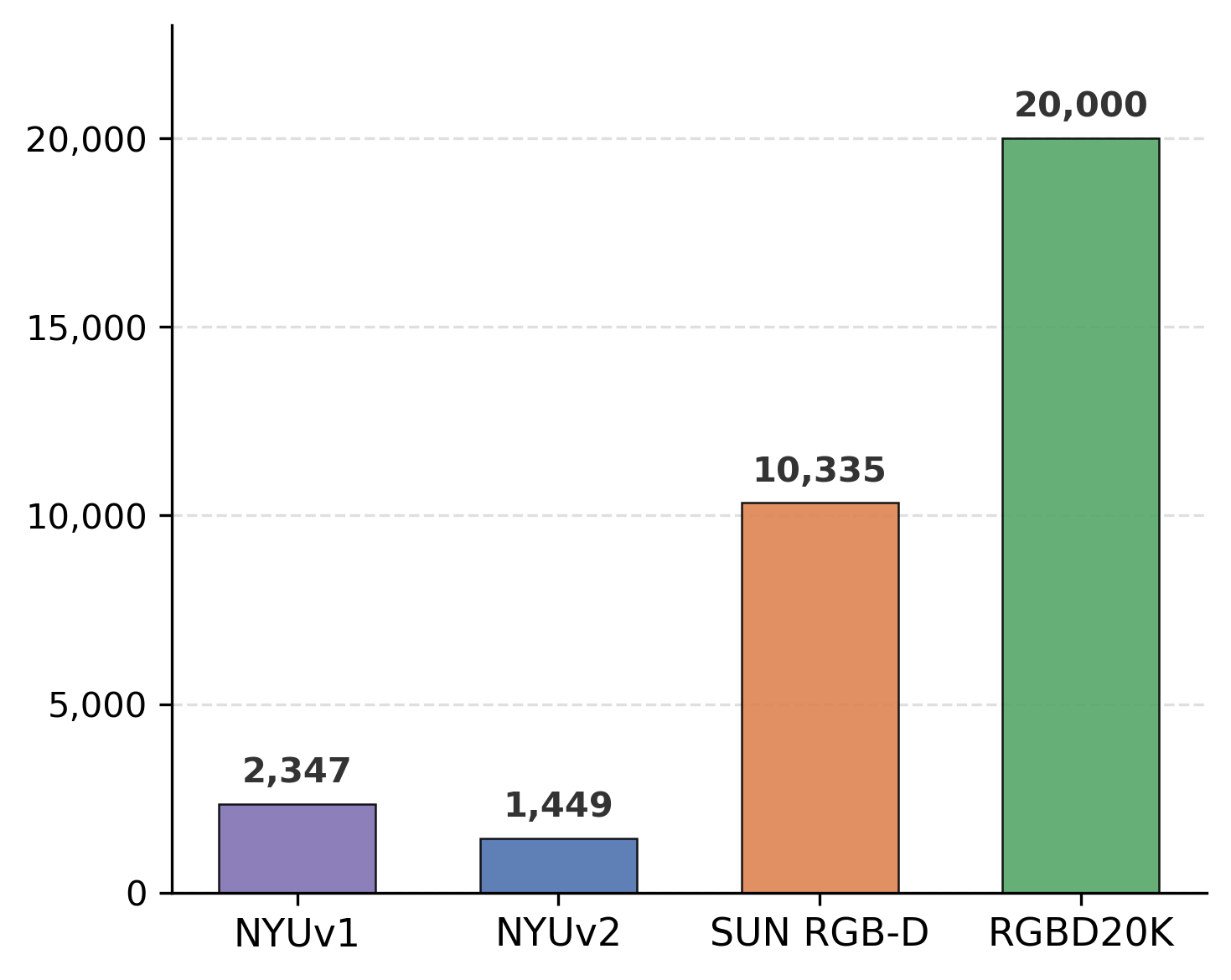}
        \caption{Number of Images}
        \label{fig:sub_num_images}
    \end{subfigure}
    \hfill
    \begin{subfigure}[b]{0.31\textwidth}
        \centering
        \includegraphics[width=\textwidth]{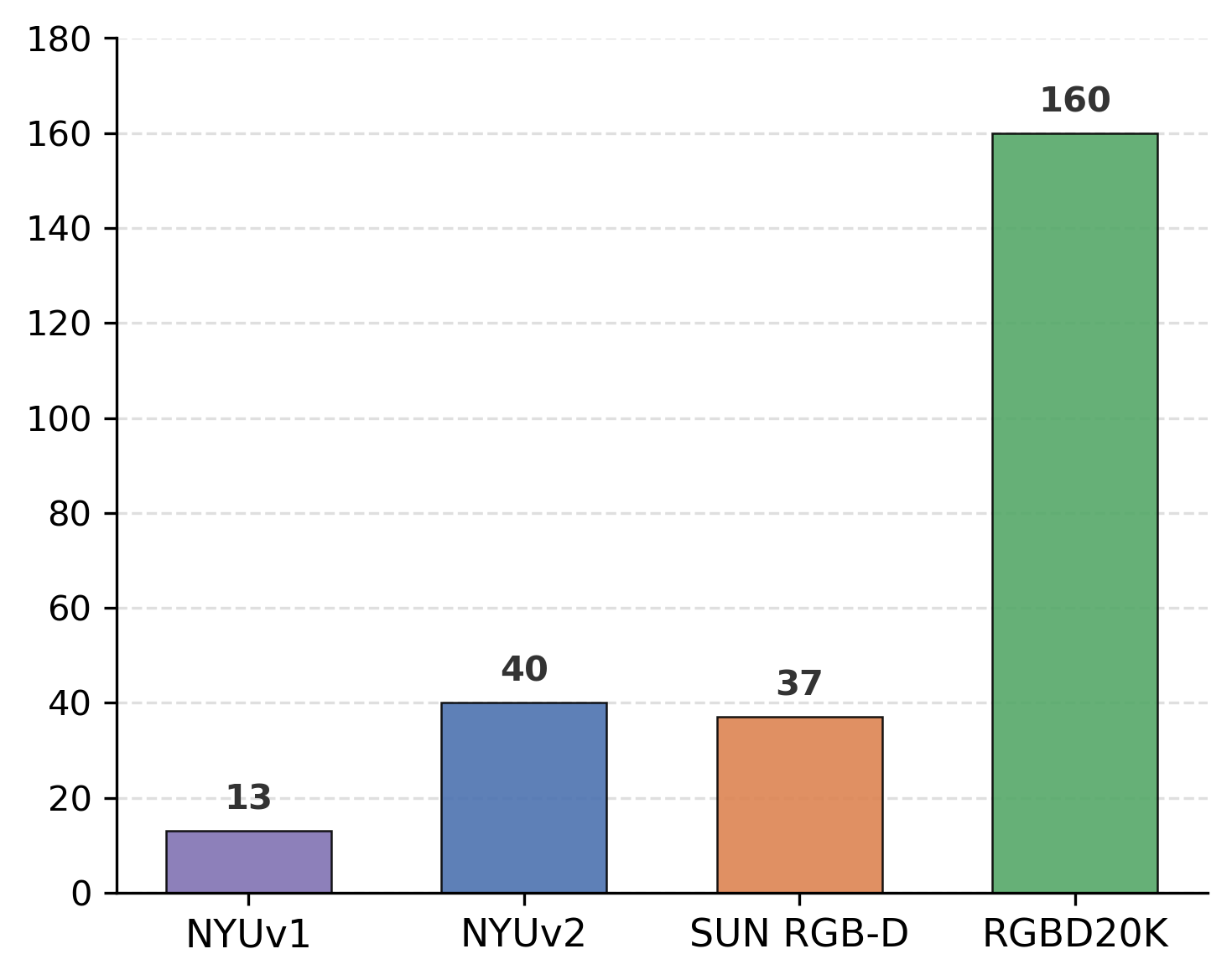}
        \caption{Object Categories}
        \label{fig:sub_obj_categories}
    \end{subfigure}
    \hfill
    \begin{subfigure}[b]{0.31\textwidth}
        \centering
        \includegraphics[width=\textwidth]{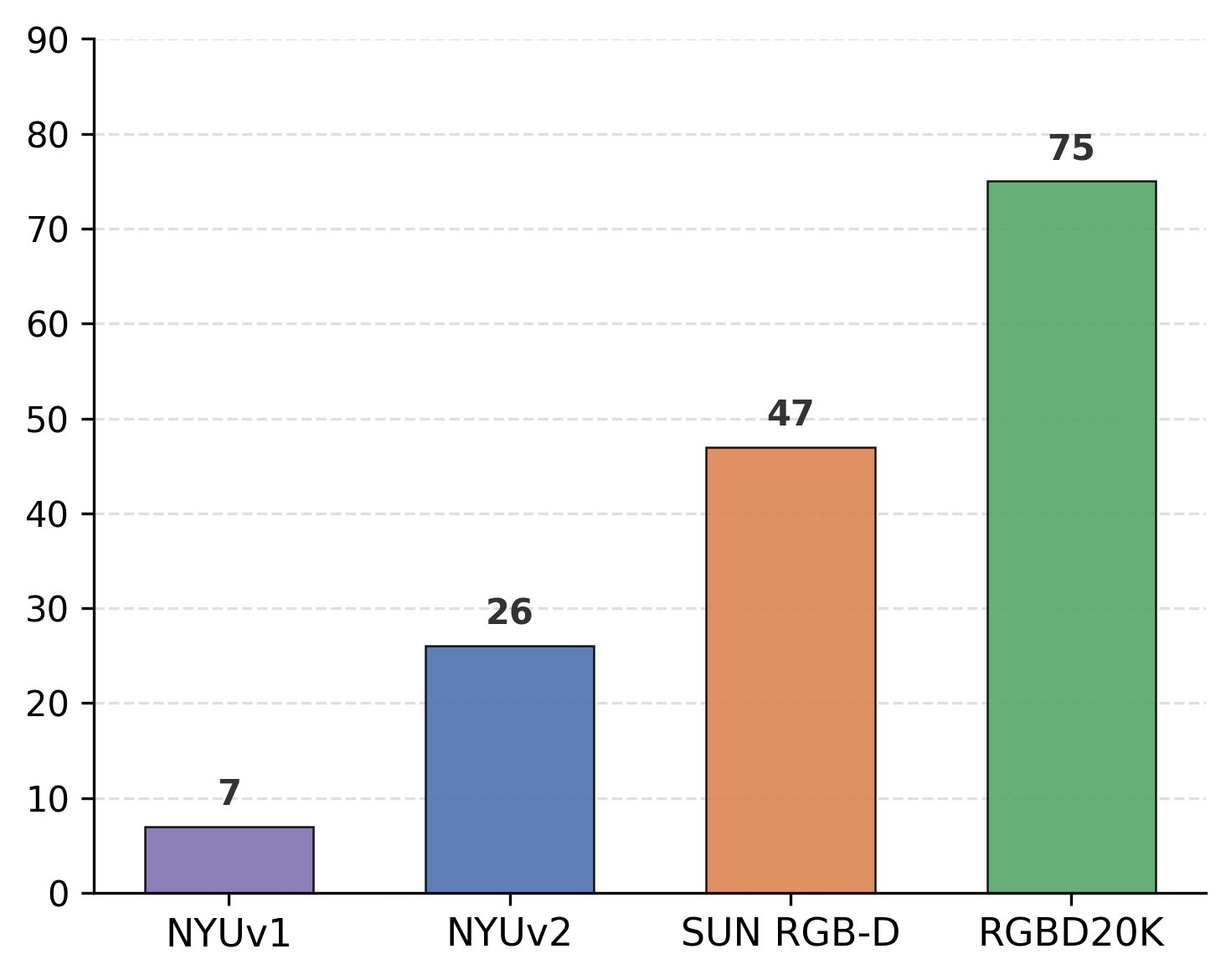}
        \caption{Scene Categories}
        \label{fig:sub_scene_categories}
    \end{subfigure}

    \caption{Comparison of the proposed RGBD20K with other RGB-D datasets.}
    \label{fig:three_images_side_by_side}
\end{figure*}

\textbf{Limited data scale.} NYUv1 \cite{silberman2011indoor} and NYUv2 \cite{silberman2012indoor} contain 2,347 and 1,449 annotated RGB-D image pairs, respectively (see Figure \ref{fig:three_images_side_by_side}(a)), and serve as early benchmarks for RGB-D semantic segmentation. However, their limited scale significantly restricts the learning capacity of modern deep models. To address this limitation, SUN RGB-D \cite{song2015sun} extends the dataset scale to 10,335 RGB-D images (see Figure \ref{fig:three_images_side_by_side} (a)). Although it significantly improves data availability and has played an important role in advancing RGB-D semantic segmentation, its scale is still insufficient for modern deep neural networks and vision transformers \cite{dosovitskiy2020image, liu2021swin}, which typically require large-scale and diverse training data to fully exploit their representation capacity and achieve strong generalization performance.

\textbf{Limited semantic coverage and scene diversity.} Beyond data scale, existing benchmarks are also constrained by limited semantic coverage and restricted scene diversity. For example, NYUv1 \cite{silberman2011indoor}, NYUv2 \cite{silberman2012indoor} and SUN RGB-D \cite{song2015sun} contain only 13, 40 and 37 semantic categories, respectively (see Figure \ref{fig:three_images_side_by_side} (b)), which are insufficient to capture the fine-grained semantic structures present in real-world environments. In addition, data collection is largely limited to relatively constrained indoor settings, leading to insufficient variation in spatial layouts, lighting conditions, occlusions, and object arrangements (see Figure \ref{fig:three_images_side_by_side} (c)). Together, these limitations in both semantic richness and environmental diversity restrict the generalization ability of models when applied to more complex and open-world scenarios.

\vspace{-0.3em}

\textbf{Limited annotation quality.}
The effectiveness of RGB-D semantic segmentation relies heavily on reliable annotations. However, existing datasets \cite{song2015sun} often suffer from imperfect or noisy labels (see Figure \ref{anno_exa_4rows}). Such deficiencies not only compromise the reliability of evaluation but also impede the development of more advanced algorithms (see Section \ref{sun rgbd analysis}). Consequently, improving annotation quality is crucial for enabling robust multimodal perception and more reliable model learning.

These limitations collectively highlight the need for a new generation of RGB-D benchmarks that provide richer semantic coverage, larger scale, and more reliable cross-modal alignment. To this end, we introduce \textbf{RGBD20K}, a large-scale benchmark designed to advance RGB-D semantic segmentation under real-world conditions. RGBD20K contains 20,000 high-quality RGB-D image pairs collected from diverse environments, with carefully refined annotations to ensure reliable supervision. Specifically, RGBD20K makes the following contributions:

\textbf{(1) \emph{Large-scale high-quality RGB-D data}.} RGBD20K contains 20,000 high-quality RGB-D image pairs, providing a significantly larger-scale benchmark compared to existing datasets.

\textbf{(2) \emph{Rich semantic coverage and diverse scene distribution}.} RGBD20K covers 160 fine-grained semantic categories, substantially exceeding existing benchmarks such as NYUv2 and SUN RGB-D, which typically contain fewer than 40 classes. In addition, the dataset spans 75 different scene types, further increasing its diversity and real-world complexity. This expanded data scale and semantic space jointly enable more comprehensive and generalizable learning of complex scene structures.

\textbf{(3) \emph{High-quality annotation refinement}.}
Unlike previous datasets that suffer from noisy labels, RGBD20K adopts a rigorous multi-stage manual refinement process to produce accurate pixel-level annotations. This process ensures clear semantic boundaries and high annotation consistency, thereby providing a more reliable supervision signal for model training.

Building upon this foundation, we further introduce the score-purified fusion (SPF) model, which follows a simple “purify-then-attend” design principle. By benefiting from the improved data quality and larger scale provided by RGBD20K, our method achieves more robust and generalizable semantic segmentation performance. By releasing RGBD20K and the SPF model, we aim to provide both a large-scale benchmark and a strong baseline to facilitate future research in robust and general-purpose RGB-D perception.

\section{Related Work}

\textbf{RGB-D Semantic Segmentation Benchmarks.} Benchmarks have been fundamental to the progression of multi-modal scene understanding. Early RGB-D benchmarks were primarily indoor-centric and designed for small-scale evaluation. NYUv2 \cite{silberman2012indoor} and SUN RGB-D \cite{song2015sun} established the initial standards, providing depth maps alongside semantic labels. However, these datasets are limited to fewer than 40 categories and often exhibit significant sensor noise and boundary misalignment. Later, ScanNet \cite{dai2017scannet} offered a larger scale of 3D indoor data, but its focus remains on voxelized reconstruction rather than high-precision 2D semantic masks. 
Matterport3D \cite{chang2017matterport3d} and 2D-3D-S \cite{armeni2017joint} introduced "building-scale" data. Matterport3D offers 194,400 RGB-D images across 90 buildings, while 2D-3D-S provides 70,496 images. Despite their massive scale, these building-level datasets were designed with different objectives. 2D-3D-S focuses on structural parsing into only 13 coarse categories (e.g., wall, floor, ceiling), while Matterport3D primarily facilitates 3D reconstruction and room-level classification. Furthermore, because these datasets are captured as continuous scans, they often contain high redundancy and "projected" labels that lack the pixel-level boundary precision. More recently, large-scale datasets such as RealSee3D \cite{Li2025realsee3d_data} have introduced 10,000 unique indoor scenes combining real-world LiDAR captures with procedurally generated environments. While RealSee3D provides an unprecedented volume of multi-view panoramic data (nearly 300,000 viewpoints), its primary focus is on 3D reconstruction, floor plan generation, and 3D detection.

Despite the emergence of such large-scale resources, there remains a critical gap in fine-grained 2D semantic perception. Many massive datasets rely on automated or coarse annotations that lack the pixel-level precision and taxonomic depth required for nuanced scene understanding. To alleviate this, our RGBD20K provides 20,000 high-fidelity image pairs with a rigorously refined 160-class taxonomy. By bridging the gap between the massive scale of modern captures like RealSee3D and the high-precision requirements of semantic segmentation, RGBD20K serves as a more challenging and reliable foundation for next-generation multimodal fusion.

\textbf{RGB-D Semantic Segmentation Algorithms.} 
RGB-D semantic segmentation \cite{zhou2022pgdenet, wang2022multimodal, dong2024efficient, cai2025keep} aims to improve recognition performance by incorporating depth information, which provides complementary 3D geometric cues that are often absent in RGB-only settings. Early mainstream approaches focused on designing complex interaction modules to fuse RGB and depth features extracted from two parallel pretrained backbones. For instance, CMX \cite{zhang2023cmx}, TokenFusion \cite{wang2022multimodal}, and GeminiFusion \cite{jia2024geminifusion} integrate multimodal representations either within the encoder or during decoding to enhance performance. However, these dual-stream architectures face two key limitations: (1) the use of separate backbones introduces significant computational overhead, and (2) initializing depth streams with RGB-pretrained weights often leads to distribution mismatch. To address these issues, recent methods such as DPLNet \cite{dong2024efficient} explore prompt-based designs to reduce the number of trainable parameters, while DFormer \cite{yin2023dformer, yin2025dformerv2} investigates unified RGB-D representation learning. By acknowledging the lower information density of depth data, DFormer allocates fewer channels to depth encoding, improving efficiency while mitigating distribution shift.

In contrast, our approach is instantiated as the score-purified fusion (SPF) network, following a simple “purify-then-attend” design principle. By performing score-based feature purification prior to cross-modal interaction, the model enables a more direct and efficient utilization of multimodal cues compared to traditional interaction-heavy or unified-backbone paradigms.

\textbf{Other Multi-modal Segmentation Benchmarks and Algorithms.} Beyond the RGB-D domain, multi-modal semantic segmentation has been extensively explored to enhance robustness in adverse environments. In the field of RGB-Thermal (RGB-T) segmentation, benchmarks such as MFNet \cite{ha2017mfnet} and PST900 \cite{shivakumar2020pst900} were introduced to address challenges in low-illumination and nighttime scenarios. Building on these, SemanticRT \cite{ji2023semanticrt} and the Multispectral Video Semantic Segmentation benchmark \cite{ji2023multispectral} have further scaled up the data volume and complexity, facilitating the development of multispectral algorithms that leverage the complementary nature of thermal and visual spectra. Recently, the DeLiVER benchmark \cite{zhang2023delivering} has pushed the boundaries of multi-modal research by providing a massive dataset covering Depth, LiDAR, multiple Views, Events, and RGB. The development of these benchmarks has driven a variety of multi-modal algorithms \cite{zhou2022edge, zhou2022mtanet, zhou2023cacfnet, dong2022gebnet, dong2023egfnet} designed to handle diverse sensing data. Early RGB-T segmenters focused on cross-modal fusion modules to align thermal and spatial features.

\begin{figure}[htbp]
    \centering
    
    \begin{subfigure}{0.48\textwidth}
        \centering
        \includegraphics[width=\linewidth]{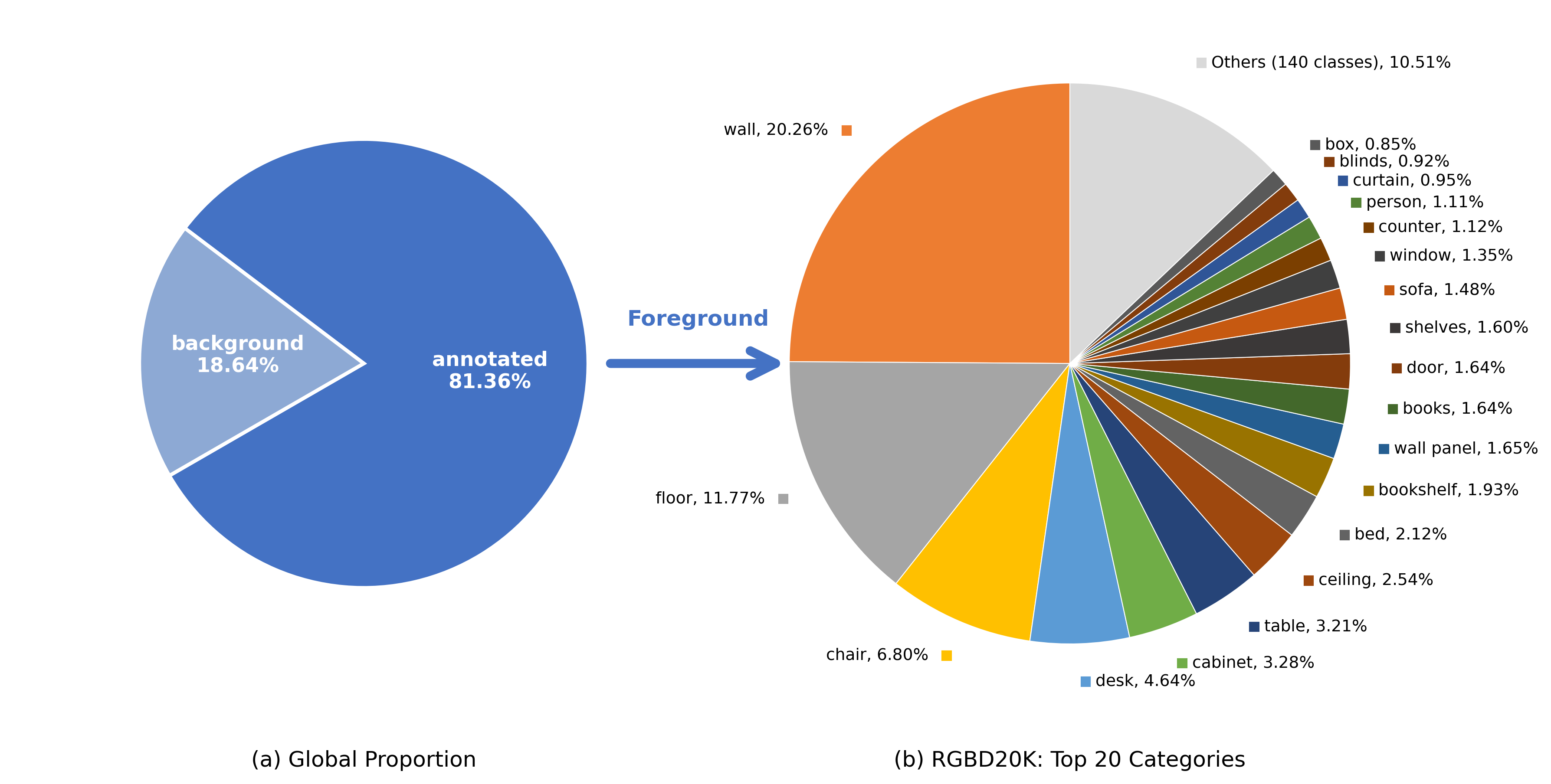}
        \caption{Pixel-level distribution}
    \end{subfigure}
    \hfill
    \begin{subfigure}{0.48\textwidth}
        \centering
        \includegraphics[width=\linewidth]{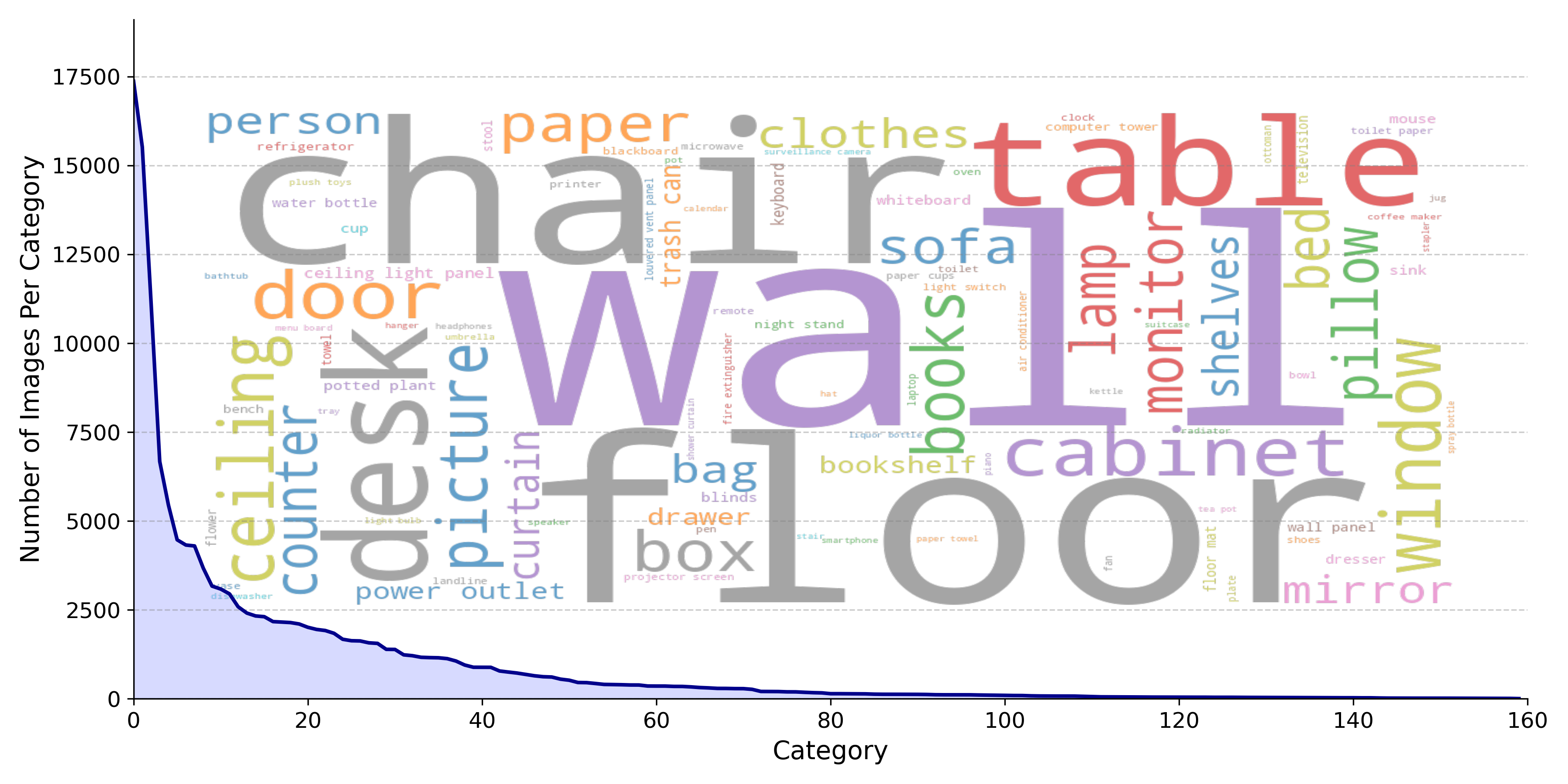}
        \caption{Category-level distribution}
    \end{subfigure}
    
    \caption{Visualization of RGBD20K distribution.}
    \label{fig:two_datasets_distribution}
\end{figure}

\begin{table}[t]
\centering
\scriptsize
\setlength{\tabcolsep}{2.5pt}
\renewcommand{\arraystretch}{0.85}
\caption{Comparison of RGB-D semantic segmentation benchmarks.}
\label{tab:dataset_comparison}
\begin{tabular}{lcccccc}
\toprule
Dataset & Year & Categories & Images & Scenarios & Masks & Avg./img \\
\midrule
NYUv1~\cite{silberman2011indoor}
& 2011 & 13 & 2,347 & 7 & -- & -- \\
NYUv2~\cite{silberman2012indoor}
& 2012 & 40 & 1,449 & 26 & 33,749 & 23.3 \\
SUN RGB-D~\cite{song2015sun}
& 2015 & 37 & 10,335 & 47 & 146,617 & 14.2 \\
RGBD20K (Ours)
& 2026 & \textbf{160} & \textbf{20,000} & \textbf{75}
& \textbf{368,312} & \textbf{18.4} \\
\bottomrule
\end{tabular}
\end{table}

\section{The Proposed RGBD20K}

\subsection{Construction Principles}

The primary goal of RGBD20K is to establish a comprehensive benchmark that provides a large-scale collection of images, rich object categories, and high-precision semantic annotations, thereby facilitating the development of more generalizable and robust RGB-D semantic segmentation methods. To this end, we follow the following principles in constructing RGBD20K:

\begin{figure}[!t]
    \centering
    \small
    \setlength{\tabcolsep}{1pt}
    \renewcommand{\arraystretch}{0.8}

    \begin{tabular}{cccc}
        \includegraphics[width=0.235\columnwidth, height=0.15\columnwidth]{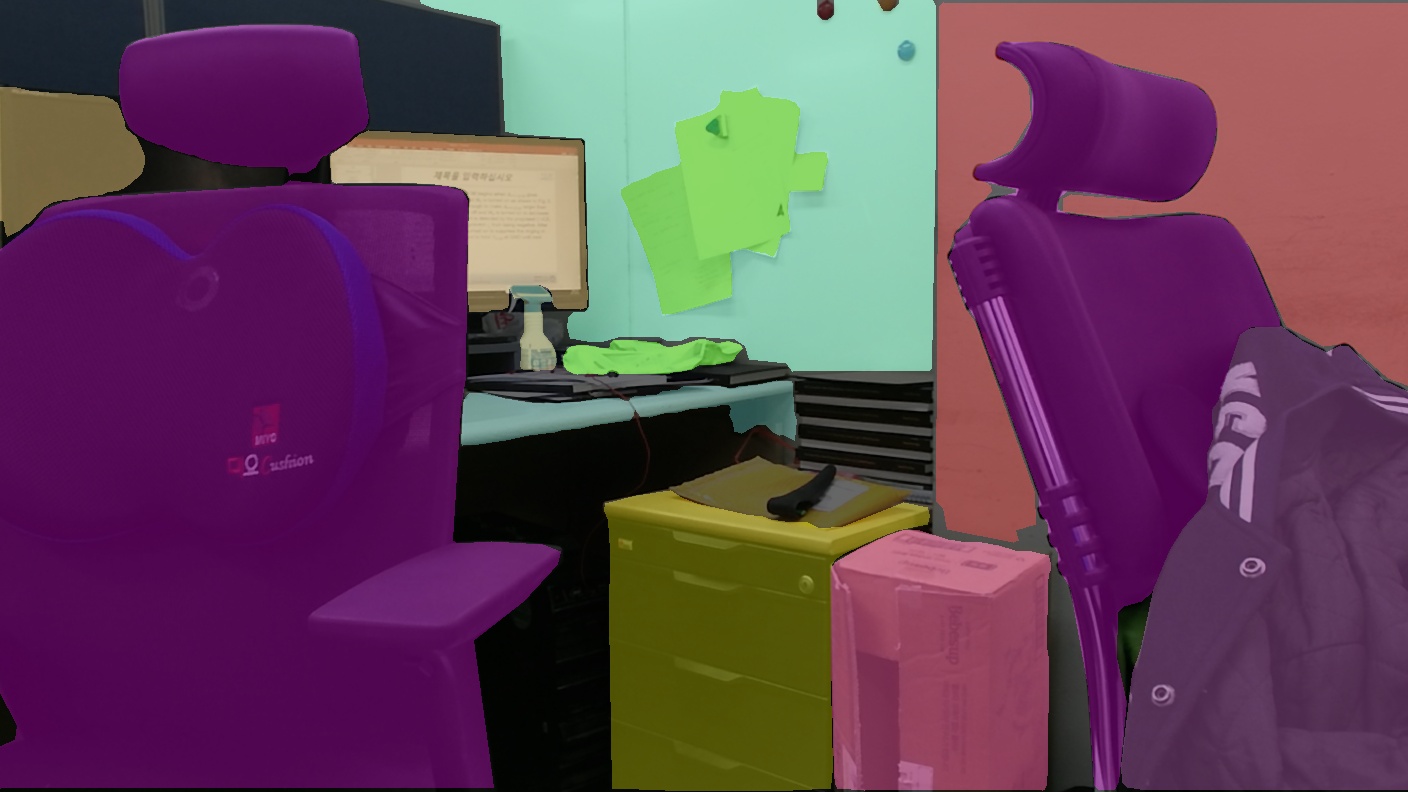} &
        \includegraphics[width=0.235\columnwidth, height=0.15\columnwidth]{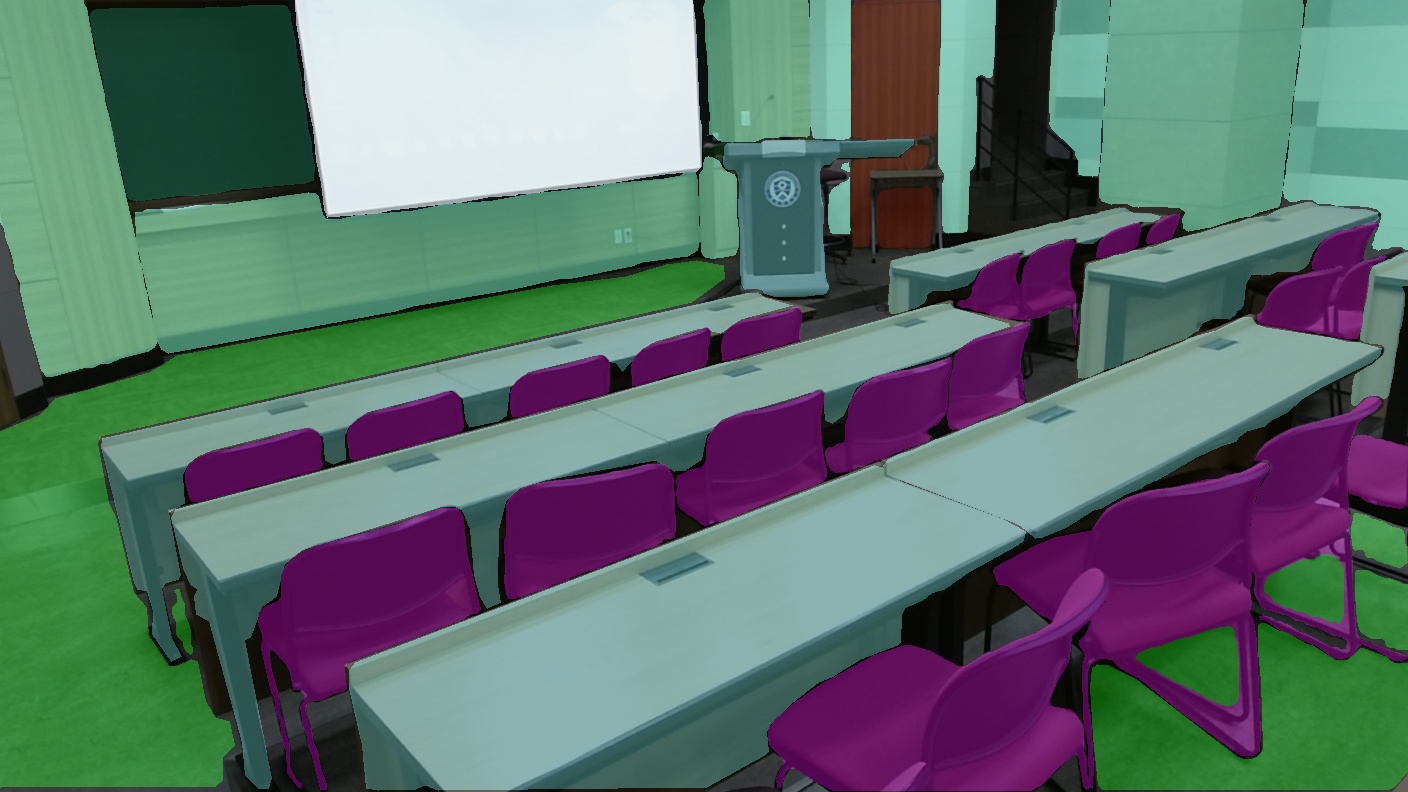} &
        \includegraphics[width=0.235\columnwidth, height=0.15\columnwidth]{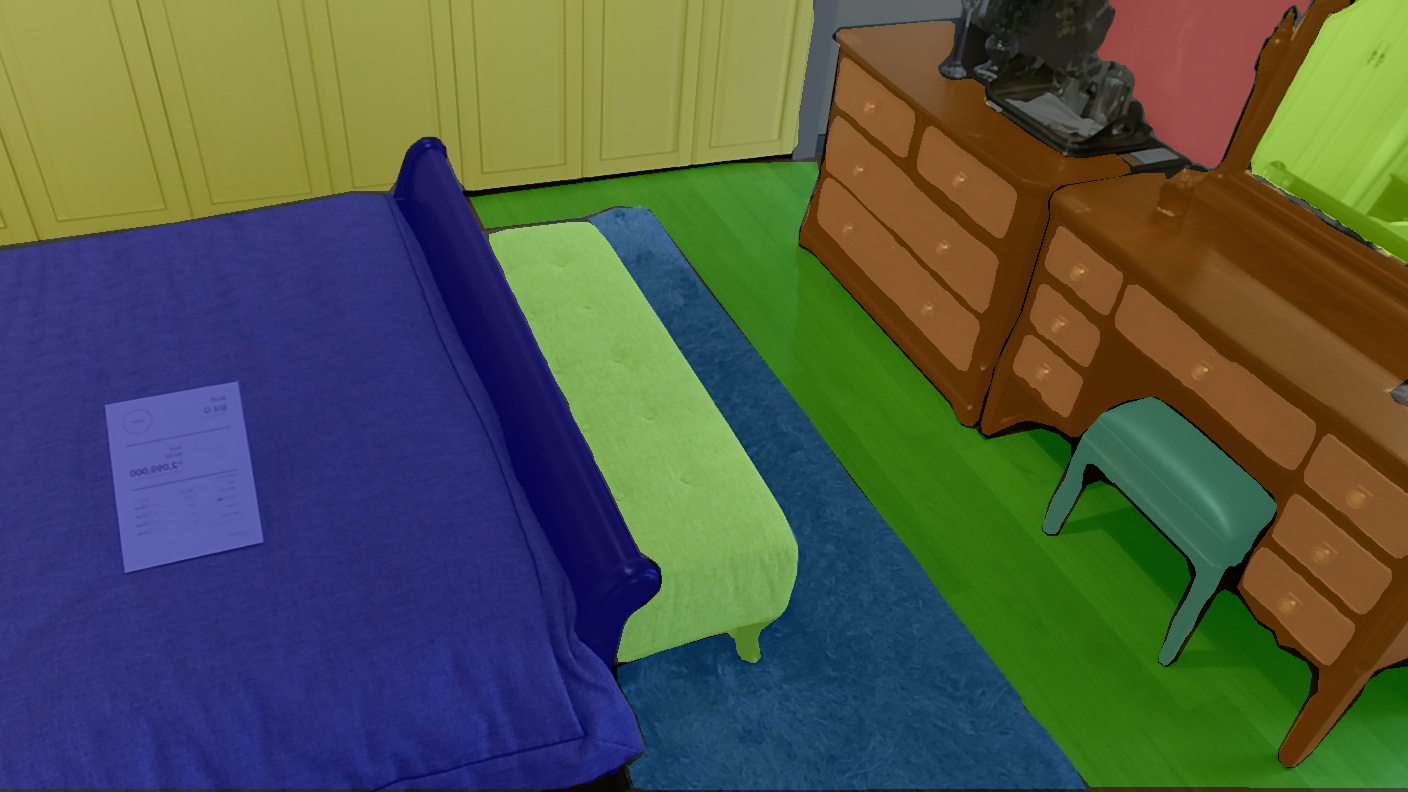} &
        \includegraphics[width=0.235\columnwidth, height=0.15\columnwidth]{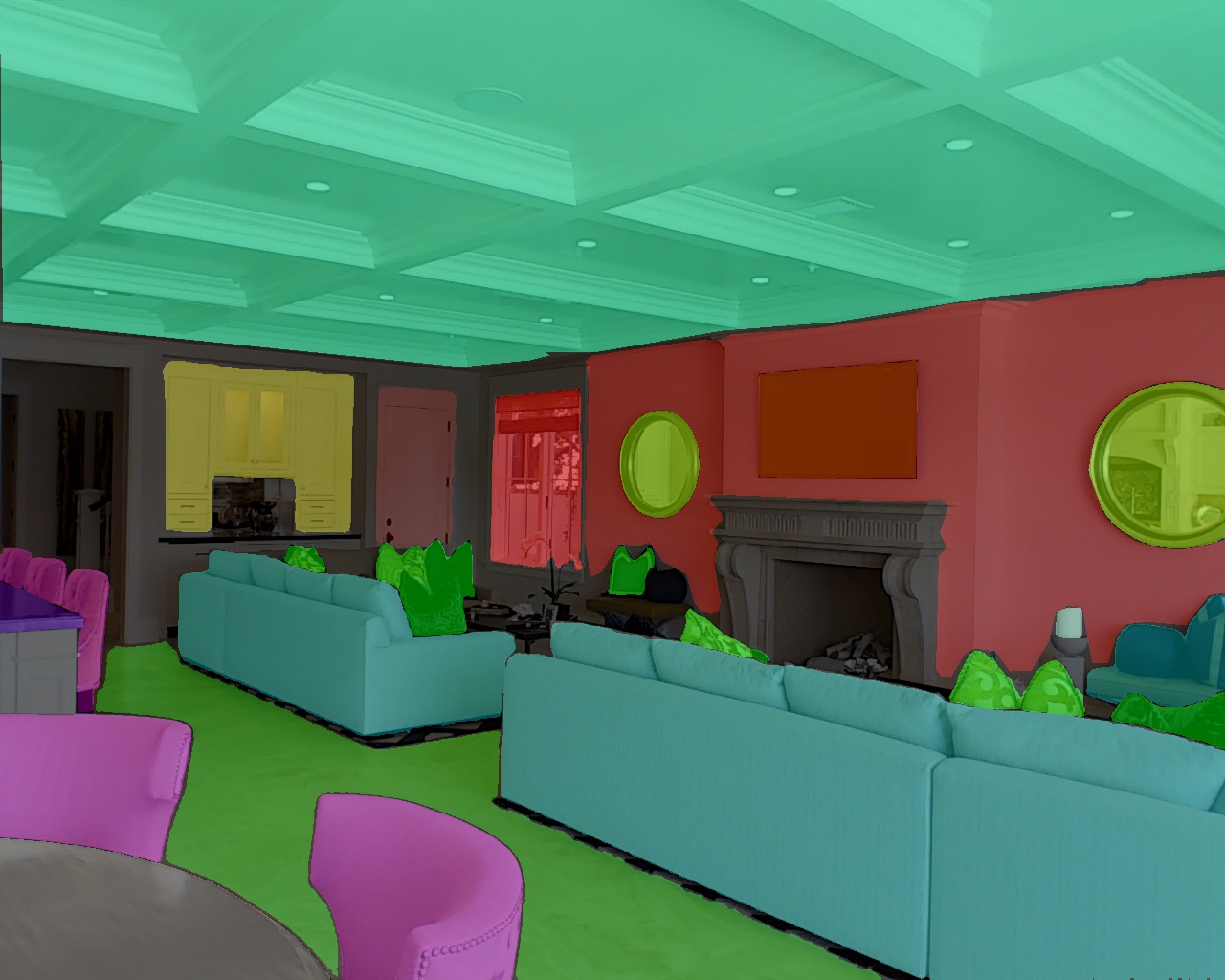} \\

        \makecell[t]{\scriptsize Office} &
        \makecell[t]{\scriptsize Classroom} &
        \makecell[t]{\scriptsize Bedroom} &
        \makecell[t]{\scriptsize Living Room} \\[\parskip]

        \includegraphics[width=0.235\columnwidth, height=0.15\columnwidth]{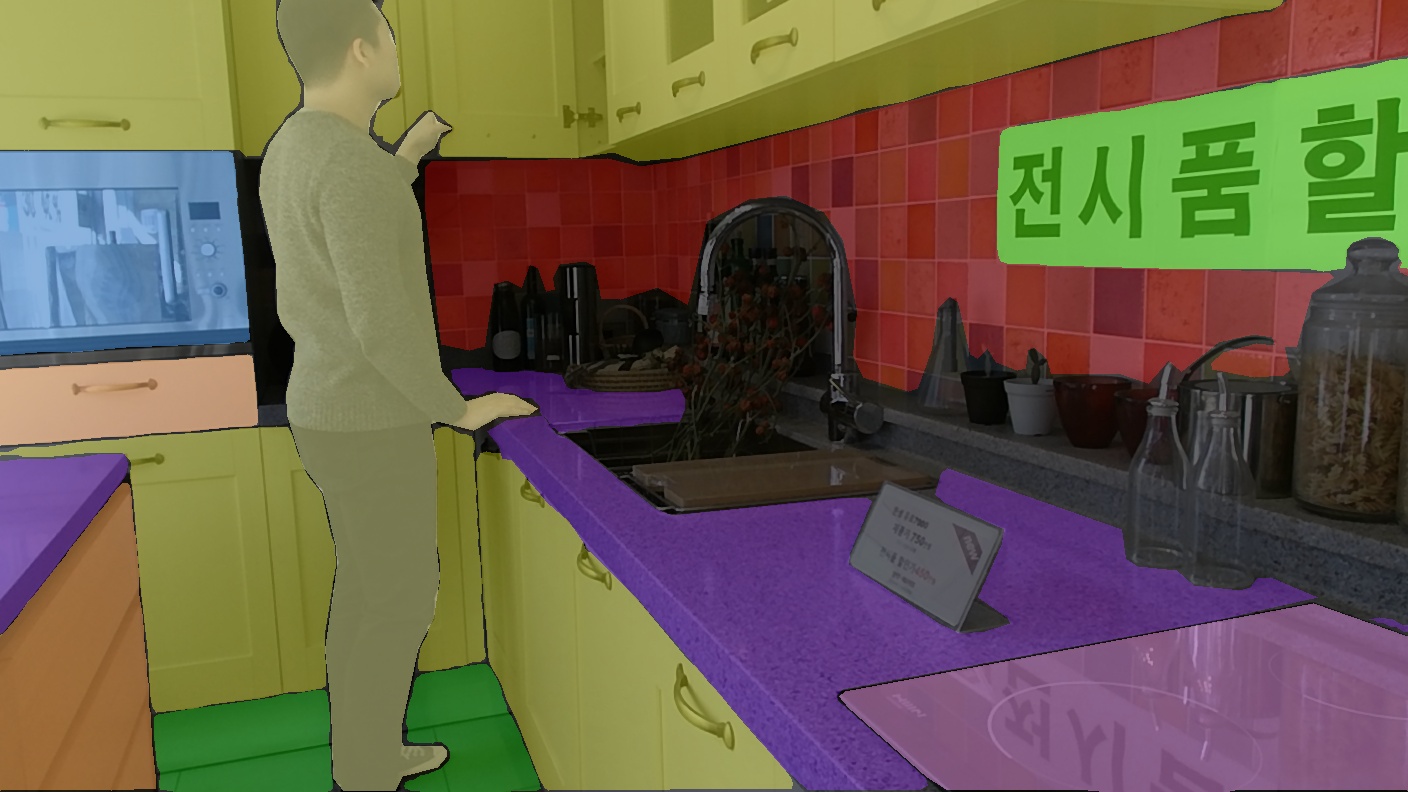} &
        \includegraphics[width=0.235\columnwidth, height=0.15\columnwidth]{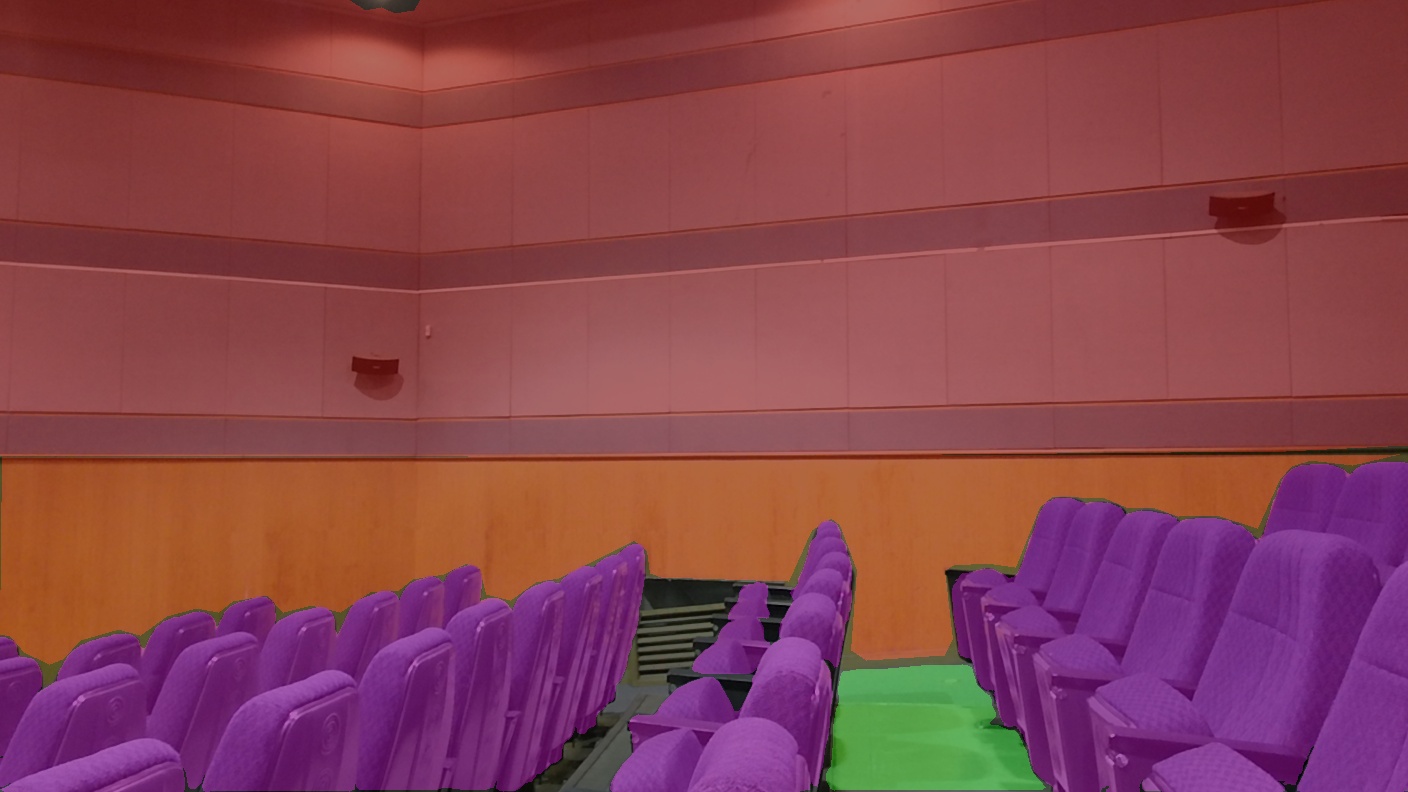} &
        \includegraphics[width=0.235\columnwidth, height=0.15\columnwidth]{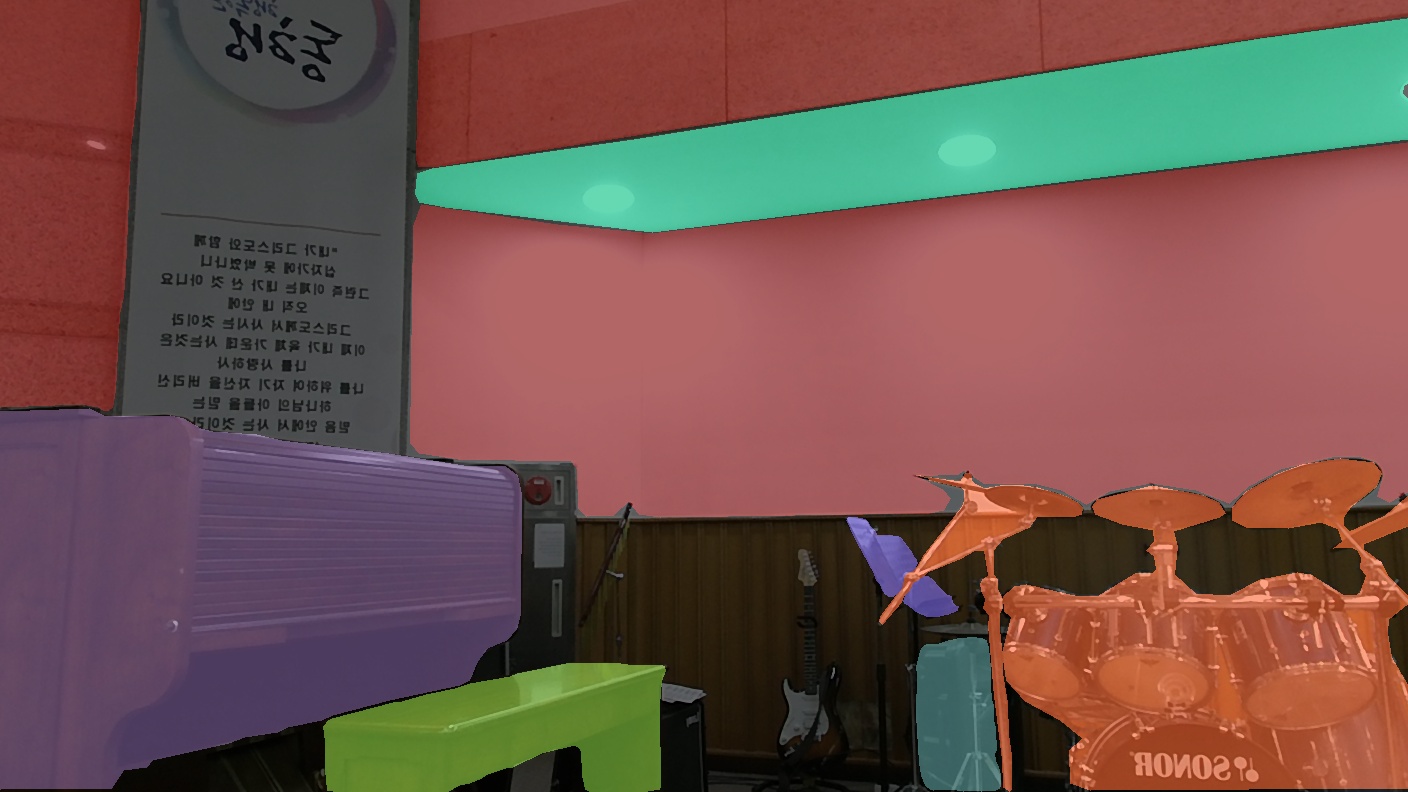} &
        \includegraphics[width=0.235\columnwidth, height=0.15\columnwidth]{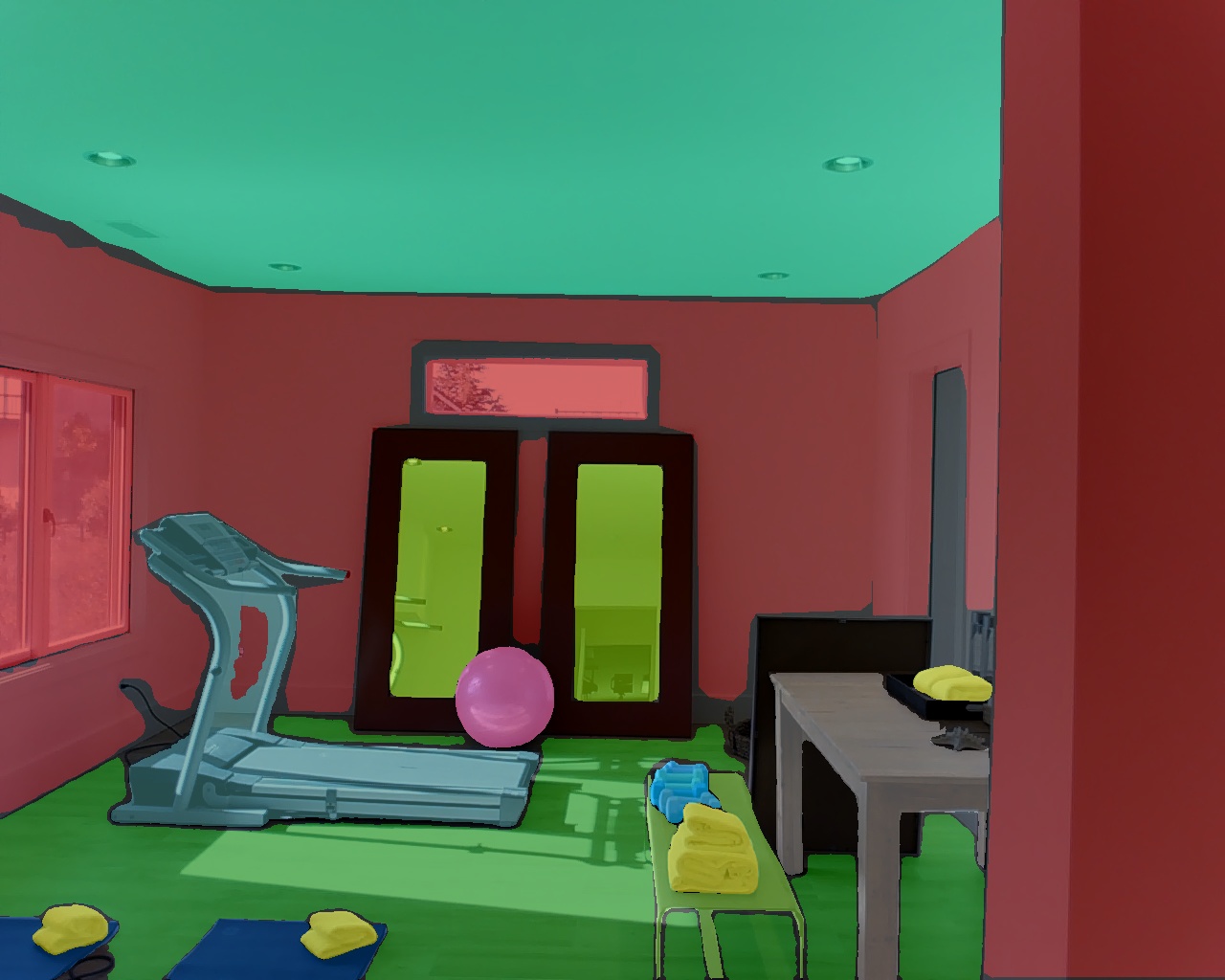} \\

        \makecell[t]{\scriptsize Kitchen} &
        \makecell[t]{\scriptsize Auditorium} &
        \makecell[t]{\scriptsize Music Room} &
        \makecell[t]{\scriptsize Gym} \\[\parskip]

        \includegraphics[width=0.235\columnwidth, height=0.15\columnwidth]{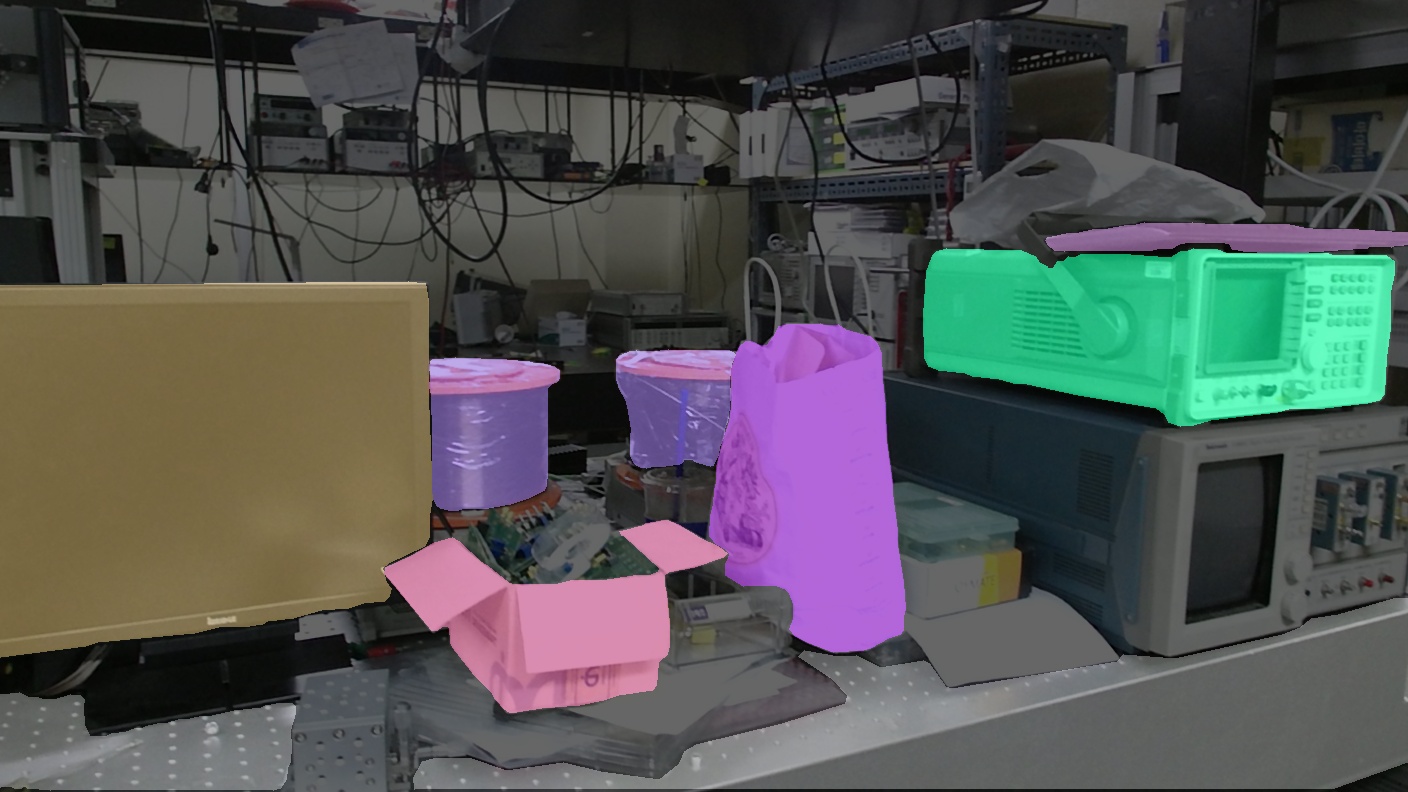} &
        \includegraphics[width=0.235\columnwidth, height=0.15\columnwidth]{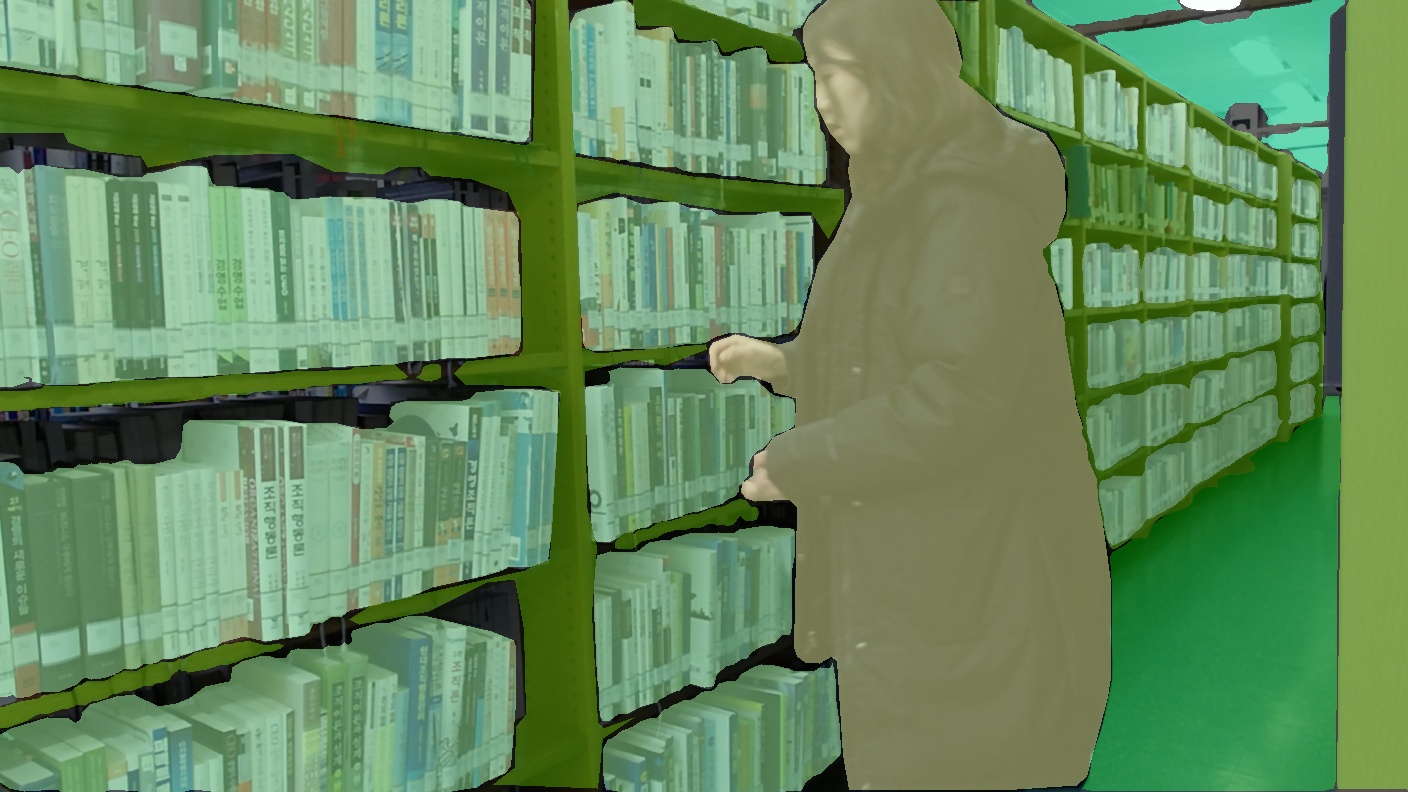} &
        \includegraphics[width=0.235\columnwidth, height=0.15\columnwidth]{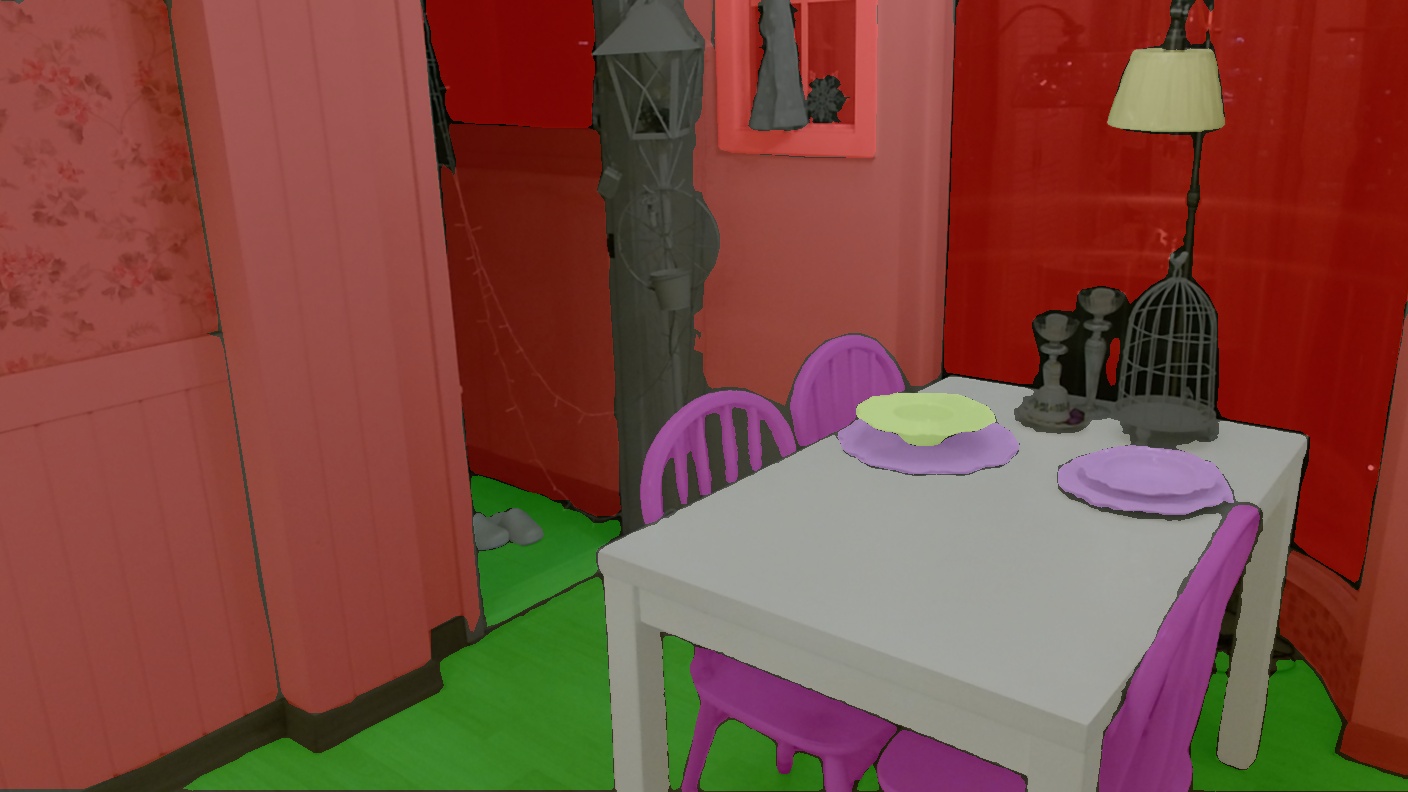} &
        \includegraphics[width=0.235\columnwidth, height=0.15\columnwidth]{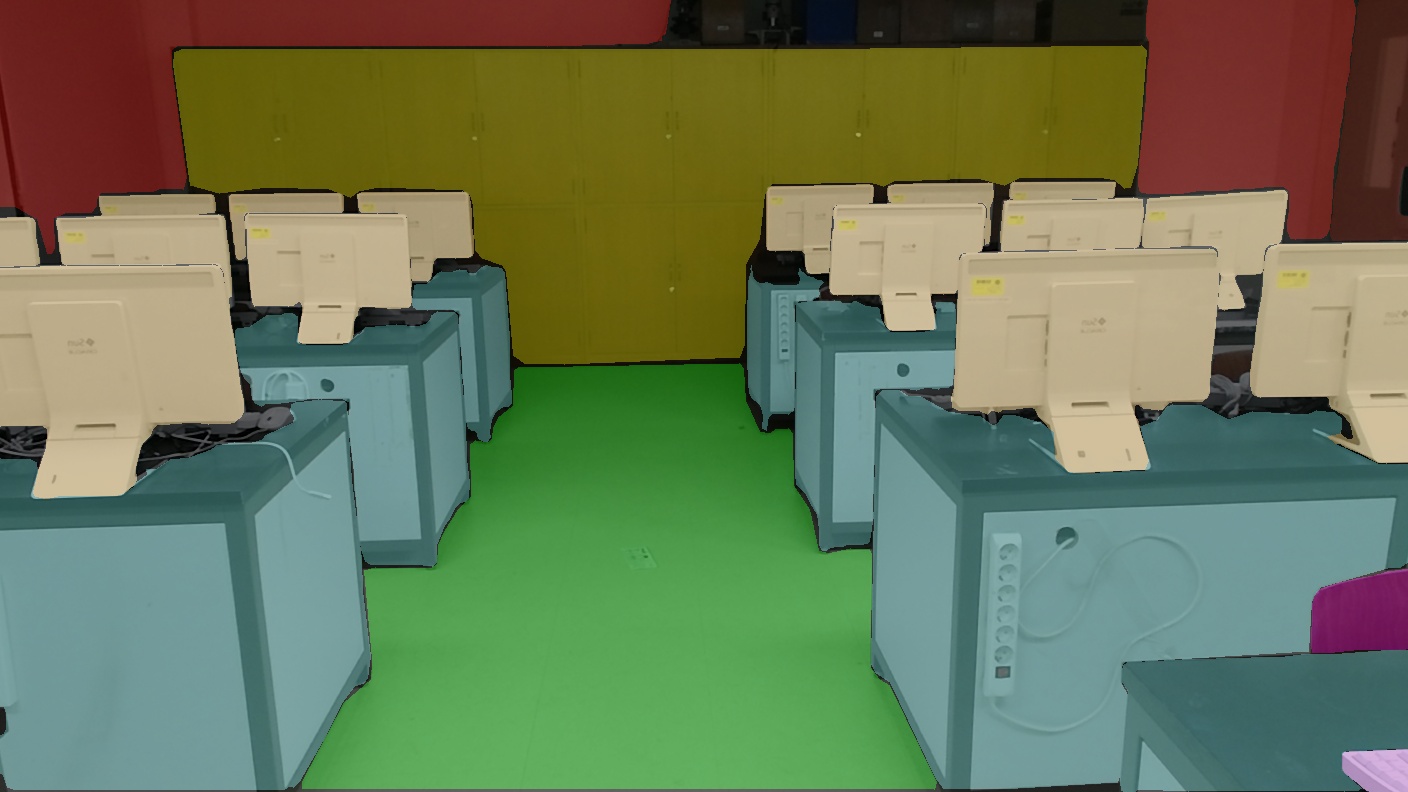} \\

        \makecell[t]{\scriptsize Lab} &
        \makecell[t]{\scriptsize Library} &
        \makecell[t]{\scriptsize Dining Room} &
        \makecell[t]{\scriptsize Computer Lab} \\[\parskip]

        \includegraphics[width=0.235\columnwidth, height=0.15\columnwidth]{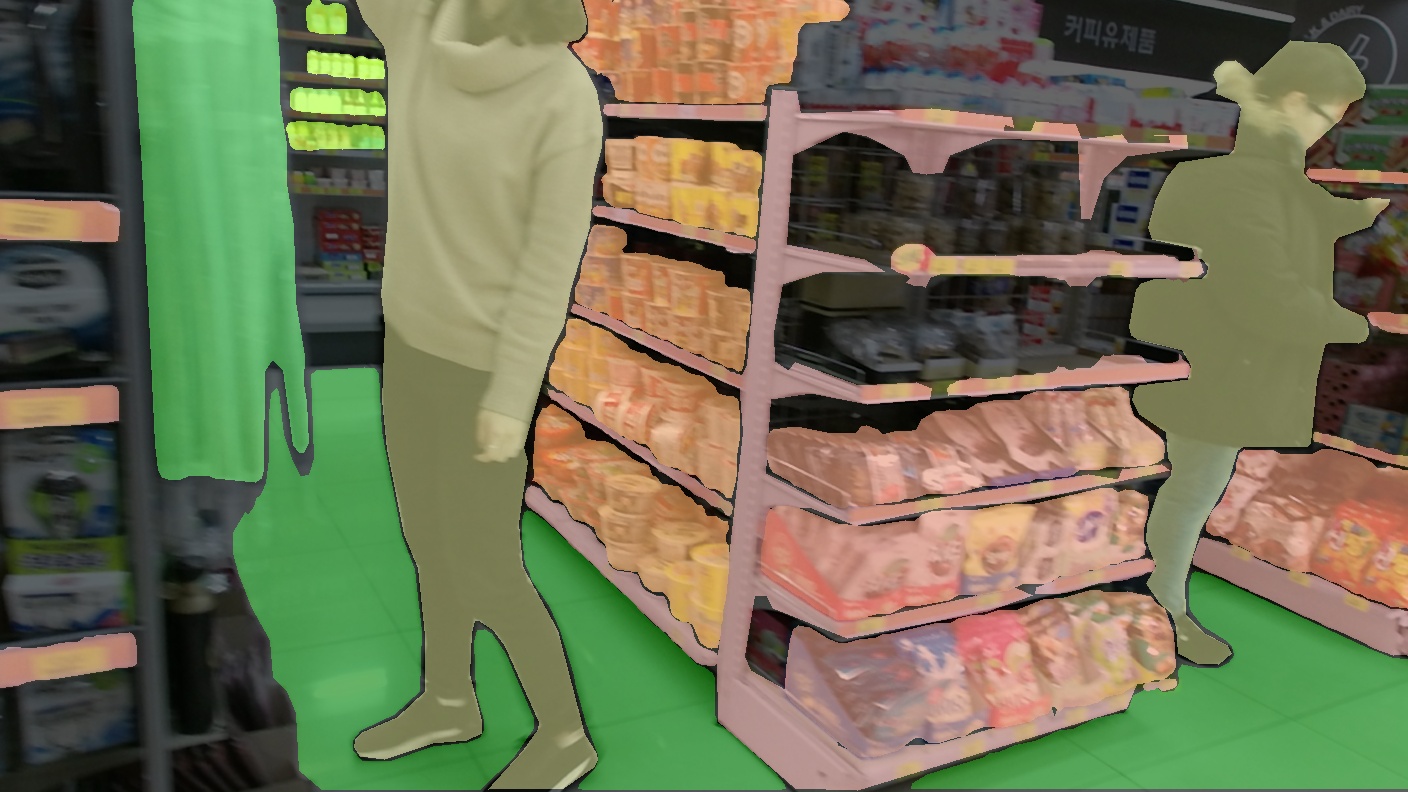} &
        \includegraphics[width=0.235\columnwidth, height=0.15\columnwidth]{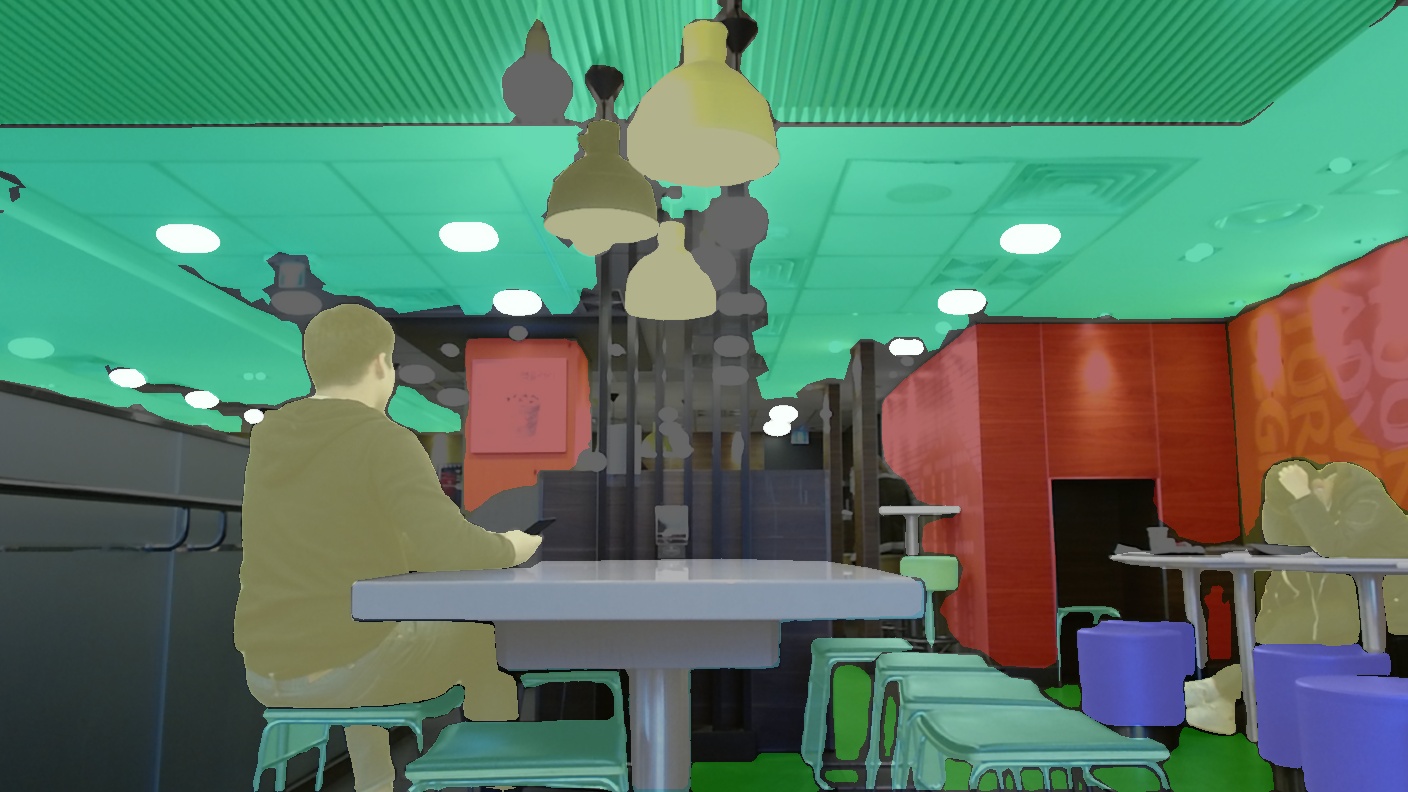} &
        \includegraphics[width=0.235\columnwidth, height=0.15\columnwidth]{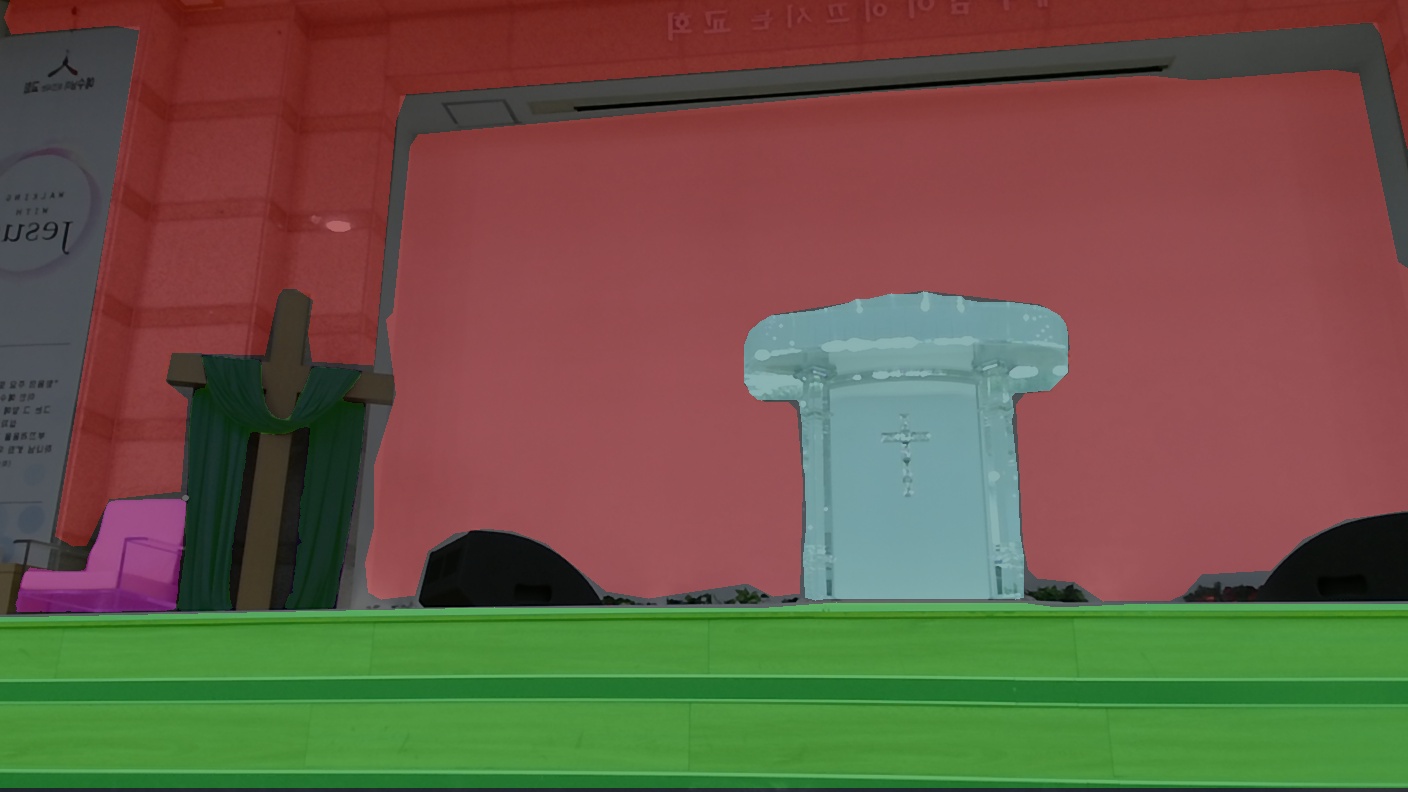} &
        \includegraphics[width=0.235\columnwidth, height=0.15\columnwidth]{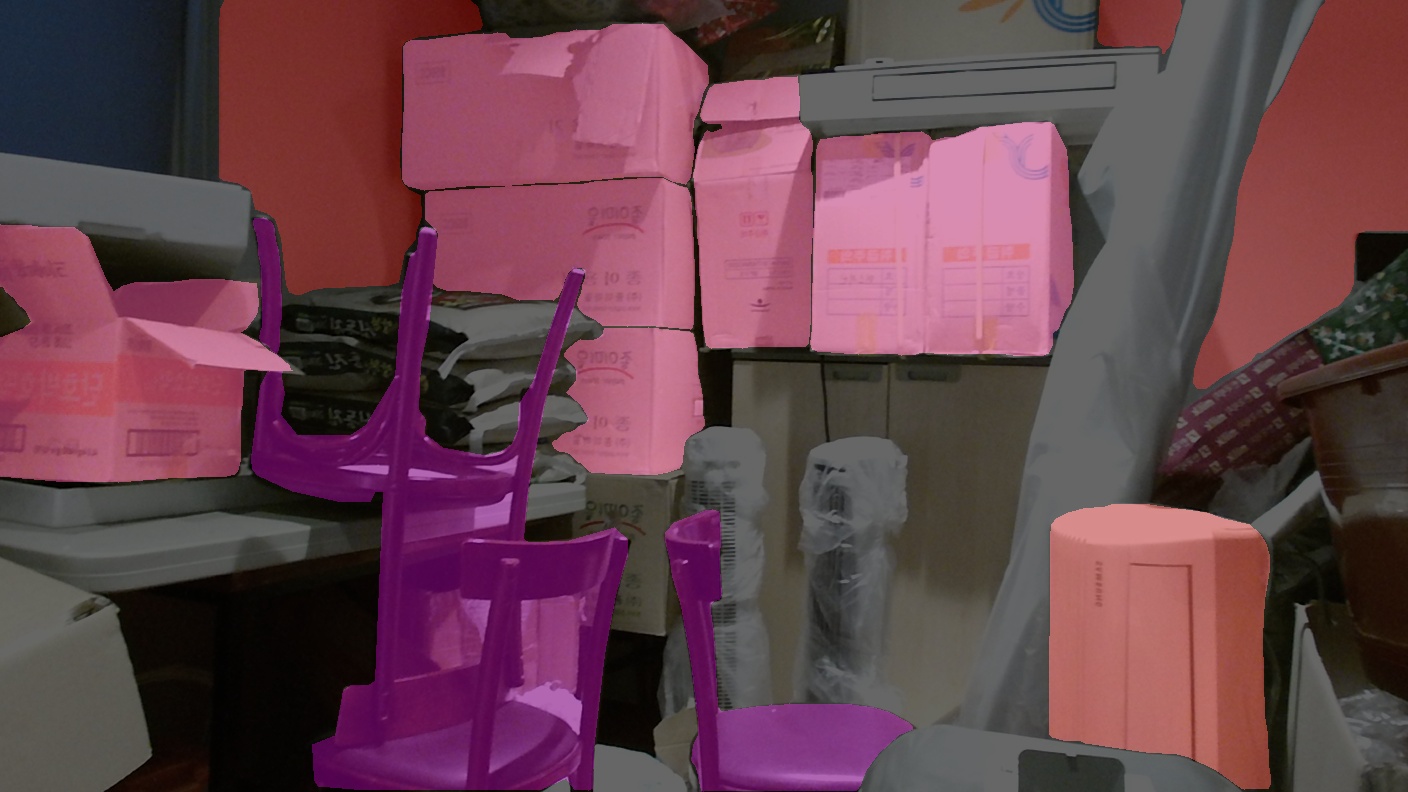} \\

        \makecell[t]{\scriptsize Retail Store} &
        \makecell[t]{\scriptsize Restaurant} &
        \makecell[t]{\scriptsize Church} &
        \makecell[t]{\scriptsize Warehouse}
    \end{tabular}

    \caption{Example images with annotations from RGBD20K.}
    \label{fig:full_grid}
    \vspace{-2mm}
\end{figure}

\textbf{Larger Scale.}
Large-scale data is crucial for training data-driven models. RGBD20K contains 20,000 RGB-D image pairs, exhibiting rich variations in illumination conditions, object arrangements, and scene layouts. Compared with existing RGB-D benchmarks, it significantly increases the data scale, providing stronger support for training more powerful segmentation models.

\textbf{Vast Categories.}
A key objective of RGBD20K is to improve category diversity and generalization ability in RGB-D semantic segmentation. To this end, the dataset includes 160 fine-grained object classes covering a wide range of common indoor objects, enabling more detailed scene understanding and semantic reasoning. In addition, it includes 75 scene types, further enhancing environmental diversity and real-world complexity.

\textbf{High-Quality Annotation.}
Annotation quality is critical for both training and evaluation in semantic segmentation. To ensure the high quality of RGBD20K, each RGB-D pair undergoes multiple rounds of manual inspection and refinement, which significantly improves boundary accuracy and annotation consistency while effectively reducing noise.

\begin{table}[t]
    \centering
    \small
    \caption{Summary of RGBD20K Dataset Collection Sources. (SS: Semantic Segmentation, SOD: Salient Object Detection, VOT: Visual Object Tracking, PT: Pre-training.)}
    \label{tab:dataset_sources}
    \begin{tabular}{lcc}
        \toprule
        Dataset & Images & Task \\
        \midrule
        SUN RGB-D \cite{song2015sun} & 10,335 & SS \\
        RGB-D mirror \cite{mei2021depth} & 824 & SS \\
        VidSOD \cite{lin2024vidsod} & 100 & SOD \\
        DepthTrack \cite{yan2021depthtrack} & 98 & VOT \\
        RGBD1K \cite{zhu2023rgbd1k} & 235 & VOT \\
        ARKitTrack \cite{zhao2023arkittrack} & 43 & VOT \\
        DIML RGB-D \cite{cho2021diml} & 8,365 & PT \\
        \midrule
        \textbf{RGBD20K} & \textbf{20,000} & \textbf{SS} \\
        \bottomrule
    \end{tabular}
\end{table}

\subsection{Construction Principles}
\subsection{Data Acquisition}
RGBD20K is built through a large-scale curation and unification process of heterogeneous RGB-D sources to ensure both environmental and semantic diversity. We collect 20,000 depth-aligned image pairs from scene-centric and tracking-oriented benchmarks (see Table \ref{tab:dataset_sources}), including SUN RGB-D \cite{song2015sun}, as well as subsets from RGB-D Mirror \cite{mei2021depth}, VidSOD \cite{lin2024vidsod}, DepthTrack \cite{yan2021depthtrack}, and ARKitTrack \cite{zhao2023arkittrack}, together with 8,365 pairs from the DIML RGB-D dataset \cite{cho2021diml}. This integration covers a broad range of real-world indoor environments and scenarios.

To ensure high fidelity, we perform a rigorous manual cleaning and re-annotation process, unifying these disparate sources under a single 160-class taxonomy. Each selected category has been verified by domain experts to ensure it is meaningful for semantic perception. The resulting dataset follows a natural long-tail distribution (see Figure \ref{fig:two_datasets_distribution}), mirroring real-world object frequencies to encourage the development of models that generalize effectively across both common and infrequent classes. Ultimately, RGBD20K offers a vastly larger and more precise semantic foundation than legacy benchmarks, facilitating research in supervised, open-vocabulary, and zero-shot perception tasks.

\subsection{Annotation}
We follow the similar principle as in \cite{zhou2019semantic, everingham2010pascal} for the semantic segmentation annotation. All images are annotated through a unified manual labeling process. Each RGB-D pair is processed by trained annotators using an interactive labeling interface, producing pixel-level semantic masks for all visible regions. A hierarchical labeling scheme is used, organizing concepts from coarse categories (e.g., furniture, appliances) to fine-grained classes (e.g., types of tables, electronic devices).

To handle occlusions in indoor scenes, we apply depth-aware ordering when constructing final masks. Objects are assigned relative depth layers from the depth map, with background regions such as walls and floors placed at the farthest level. For overlaps, depth cues and mask geometry are used to determine consistent ordering, ensuring correct foreground–background relationships.

Unlike fixed-label benchmarks, RGBD20K supports flexible category refinement, allowing new semantic classes to be added during annotation for better coverage of real-world concepts. All regions are labeled at the semantic level to support segmentation and scene understanding.

Object parts are also annotated when applicable and linked to their parent objects, forming a lightweight hierarchical structure that reflects real-world composition (e.g., drawer–cabinet).  Figure \ref{fig:full_grid} displays several annotation examples.

\subsection{Dataset Split}

RGBD20K consists of 20,000 RGB-D image pairs collected from diverse indoor environments. We adopt a standard benchmark split for training and evaluation, using 18,000 pairs for training and 2,000 pairs for testing. The split is performed in a stratified manner to preserve the distributions of scene types, object categories, and depth characteristics across both subsets. All 160 semantic categories are included in both training and testing sets, while maintaining a long-tailed distribution consistent with real-world indoor scenes. Although the test set accounts for only 10\% of the data, it is designed to be representative of the full dataset while enabling efficient evaluation. This split follows common practice in large-scale indoor vision benchmarks, where a compact but diverse test set is used to balance efficiency and robustness.

\section{Methodology: Score-Purified Fusion Model}

In RGB-D semantic segmentation, effectively fusing complementary information from heterogeneous modalities remains a challenging problem. Existing fusion methods, particularly those based on standard cross-attention, often suffer from attention dilution. This issue arises because the attention mechanism must simultaneously handle cross-modal inconsistencies (e.g., sensor noise and misaligned depth boundaries) while aggregating long-range contextual information, which can weaken discriminative feature learning. To address this problem, we propose the score-purified fusion (SPF) Network, following a simple “purify-then-attend” design principle. Instead of directly applying attention on raw projected features, SPF explicitly filters and refines the Key ($K$) and Value ($V$) representations at the linear projection stage before attention computation. Specifically, we introduce cross-examined reliability scores to assess feature consistency across modalities, enabling adaptive suppression of unreliable responses and enhancement of semantically consistent regions.

\begin{figure*}[!t]
    \centering
    \includegraphics[width=0.9\textwidth]{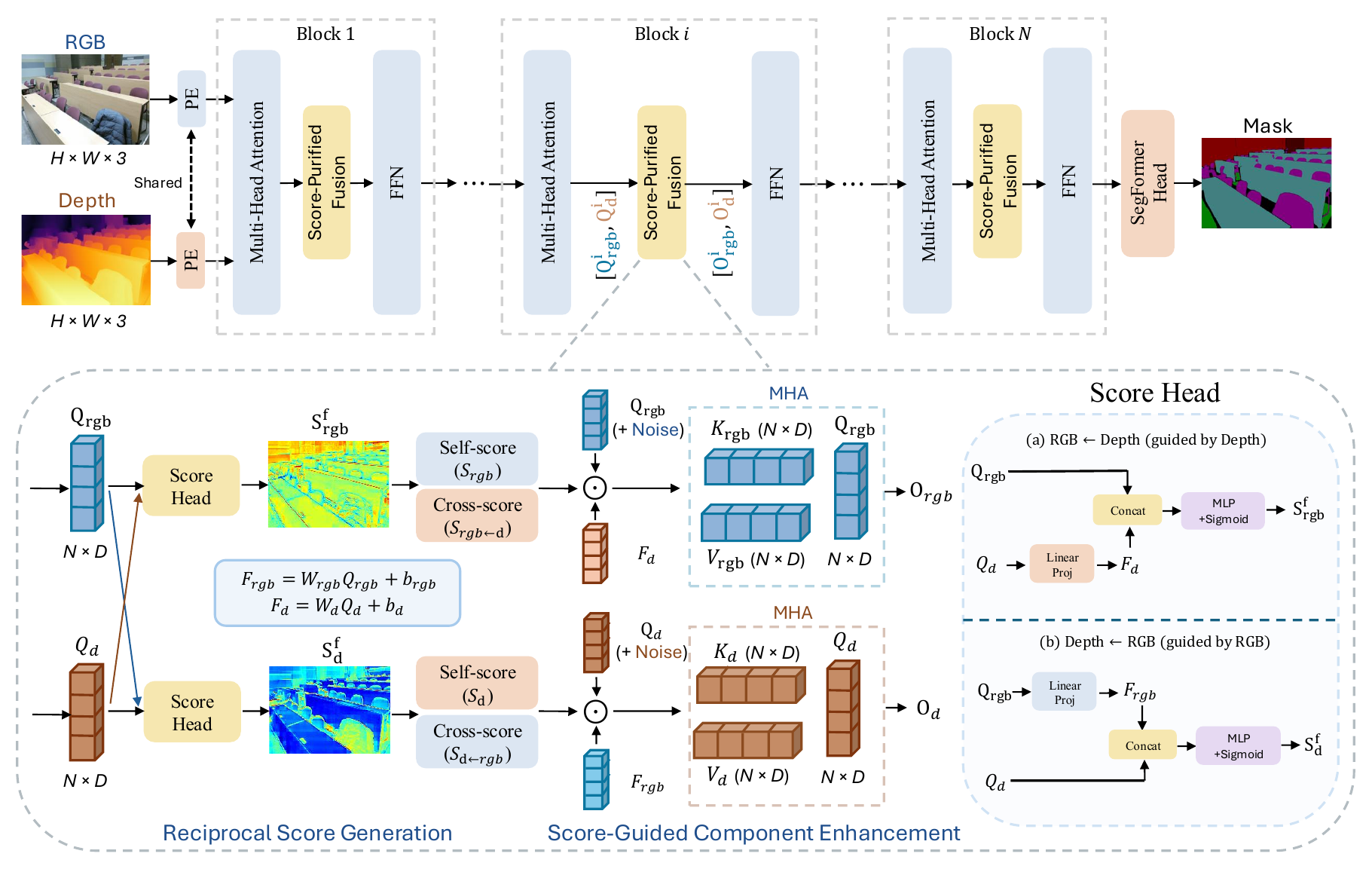}
    \caption{The overall architecture of the proposed score-purified fusion model.}
    \label{fig4}
\end{figure*}

\vspace{-0.5em}

\subsection{Overall architecture}
Our SPF model follows the GeminiFusion method \cite{jia2024geminifusion}, featuring a four-stage hierarchical encoder similar to SegFormer \cite{xie2021segformer}. As illustrated in Fig.~\ref{fig4}, the network takes RGB and Depth images as inputs. Each modality is processed through shared encoder layers, which comprises Multi-Head Attention (MHA) and Feed-Forward Network (FFN) blocks to extract multi-scale features, which are then fused at each stage. For conciseness, Fig.~\ref{fig4} illustrates only the transformer blocks within the first stage rather than depicting all four hierarchical stages in detail. Different from GeminiFusion \cite{jia2024geminifusion}, our key contribution lies in the proposed Score-Purified Fusion module, which replaces the original fusion strategy for more effective multimodal feature integration. Finally, the fused features are passed to a SegFormer head decoder to produce the segmentation predictions.

\subsection{Score-Purified Fusion}
The core of the SPF module lies in Reciprocal Score Generation and Score-Guided Manifold Purification. As shown in Fig.~\ref{fig4}, for simplicity, we omit the block index i in the following formulations and present the operations at a representative layer without loss of generality.

\paragraph{Reciprocal Score Generation.} 
Let $Q_{rgb}, Q_{d} \in \mathbb{R}^{N \times D}$ be the outputs of the Multi-Head Attention. To estimate the dynamic reliability scores of features from each modality, we introduce the \textbf{Score Head}. We first project the raw features into aligned embeddings to enable cross-modal comparison within a balanced representation space:
{\small
\begin{equation}
F_{rgb} = W_{rgb}Q_{rgb} + b_{rgb}, \quad F_{d} = W_{d}Q_{d} + b_{d}
\end{equation}
}

where $W_{rgb}, W_d \in \mathbb{R}^{D \times D}$ and $b_{rgb}, b_d \in \mathbb{R}^D$ are learnable projection parameters. Rather than applying a simple heuristic fusion, a \textbf{Relation Arbiter} is introduced to perform a fine-grained cross-examination. We estimate the fused importance scores $S^f$ that capture the pixel-wise reliability of each stream:
{\small
\begin{equation}
S^f_{rgb}=\sigma\!\left(\mathrm{MLP}([Q_{rgb};F_d])\right),\quad
S^f_d=\sigma\!\left(\mathrm{MLP}([Q_d;F_{rgb}])\right)
\end{equation}
}

where $[ \; ]$ denotes channel-wise concatenation and $\sigma$ is the Sigmoid function. These scores are subsequently decomposed into intra-modal ($\mathcal{S}_{rgb}, \mathcal{S}_{d}$) and cross-modal ($\mathcal{S}_{rgb \leftarrow d}, \mathcal{S}_{d \leftarrow rgb}$) components via a $\mathrm{Split}$ operation.

{\small
\begin{equation}
S_{rgb},\, S_{rgb \leftarrow d} = \mathrm{Split}(S^f_{rgb}), \quad 
S_{d}, S_{d \leftarrow rgb} = \mathrm{Split}(S^f_{d})
\end{equation}
}

\paragraph{Score-Guided Component Enhancement.} 
The critical innovation of our approach is the construction of purified Keys ($K$) and Values ($V$). By adaptively weighting the learnable noise $e$ and the cross-modal features $F$ by their respective reliability scores, we perform a \textbf{Purified Alignment}:

{\small
\begin{equation}
\begin{aligned}
K_{rgb} &=
\frac{(Q_{rgb}+e_{rgb}^{k})\cdot\mathcal{S}_{rgb}
+F_d\cdot\mathcal{S}_{rgb\leftarrow d}}
{\mathcal{S}_{rgb}+\mathcal{S}_{rgb\leftarrow d}+\epsilon},\\
V_{rgb} &=
\frac{(Q_{rgb}+e_{rgb}^{v})\cdot\mathcal{S}_{rgb}
+F_d\cdot\mathcal{S}_{rgb\leftarrow d}}
{\mathcal{S}_{rgb}+\mathcal{S}_{rgb\leftarrow d}+\epsilon}.
\end{aligned}
\end{equation}
}

{\small
\begin{equation}
\begin{aligned}
K_{d}&=\frac{(Q_{d}+e_{d}^{k})\cdot\mathcal{S}_{d}
+F_{rgb}\cdot\mathcal{S}_{d\leftarrow rgb}}
{\mathcal{S}_{d}+\mathcal{S}_{d\leftarrow rgb}+\epsilon},\\
V_{d}&=\frac{(Q_{d}+e_{d}^{v})\cdot\mathcal{S}_{d}
+F_{rgb}\cdot\mathcal{S}_{d\leftarrow rgb}}
{\mathcal{S}_{d}+\mathcal{S}_{d\leftarrow rgb}+\epsilon}.
\end{aligned}
\end{equation}
}

where $\cdot$ denotes the Hadamard product, $e^k, e^v \in \mathbb{R}^{N \times D}$ are learnable noise components capturing modality-specific uncertainty, and $\epsilon$ is a stability constant. This mathematical formulation allows the model to selectively filter out cross-modal noise (e.g., depth edge artifacts) before the  attention mechanism is invoked, preventing attention dilution.

With the purified Key ($K$) and Value ($V$) manifolds established, standard Multi-Head Attention (MHA) operates on a noise-robust latent space. This eliminates the burden of noise resolution from the attention mechanism, allowing it to focus entirely on high-fidelity context aggregation:

{\small
\begin{equation}
\begin{aligned}
\text{O}_{rgb} &= \text{MHA}(Q_{rgb}, K_{rgb}, V_{rgb}), \\
\text{O}_{d}   &= \text{MHA}(Q_{d}, K_{d}, V_{d}).
\end{aligned}
\end{equation}
}

The output features are then added back to the original identity streams via residual skip connections, followed by the Feed-Forward Network (FFN) within the Transformer block.

\section{Experiments}

\subsection{Datasets and Implementation Details}
\paragraph{Datasets and Metrics} 
To comprehensively evaluate our multimodal semantic segmentation method, we conduct experiments on three widely adopted benchmarks: NYUv2~\cite{silberman2012indoor}, SUN RGB-D~\cite{song2015sun}, and our newly proposed RGBD20K dataset, which together cover diverse indoor scenes and object categories, enabling a thorough assessment of model generalization across different data scales and complexities. Specifically, NYUv2 contains 795 training images and 654 testing images across 40 semantic categories, and all inputs are processed at a resolution of $480 \times 640$ following GeminiFusion~\cite{jia2024geminifusion} for fair comparison. SUN RGB-D includes 5,285 training images and 5,050 testing images across 37 categories, making it approximately $7\times$ larger than NYUv2, and uses an input resolution of $480 \times 480$ for evaluation. Lastly, our RGBD20K dataset consists of 18,000 training images and 2,000 testing images spanning 160 categories, and is evaluated with an input resolution of $480 \times 640$, providing more fine-grained annotations and significantly increasing task difficulty. Following standard evaluation protocols, we report mean Intersection-over-Union (mIoU), computed as the average IoU across all semantic categories.

\begin{figure}[t]
    \centering

    \begin{tabular}{@{}c@{\hspace{12pt}}c@{}}
        \includegraphics[width=0.11\textwidth]{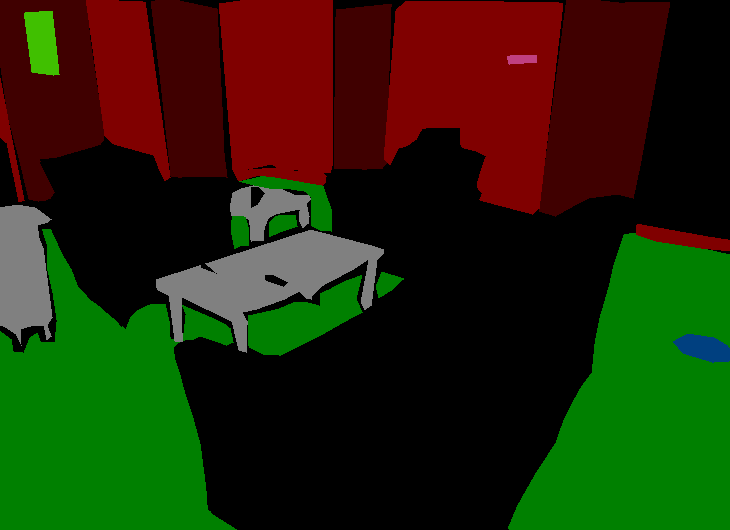}
        \includegraphics[width=0.11\textwidth]{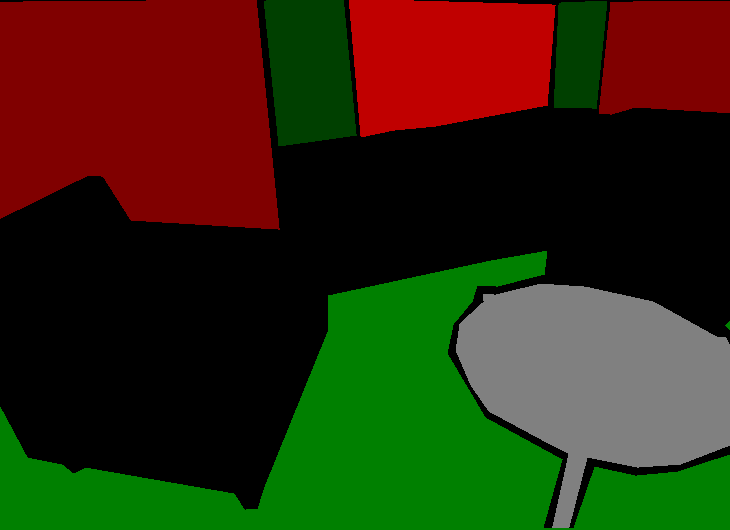} &
        \includegraphics[width=0.11\textwidth]{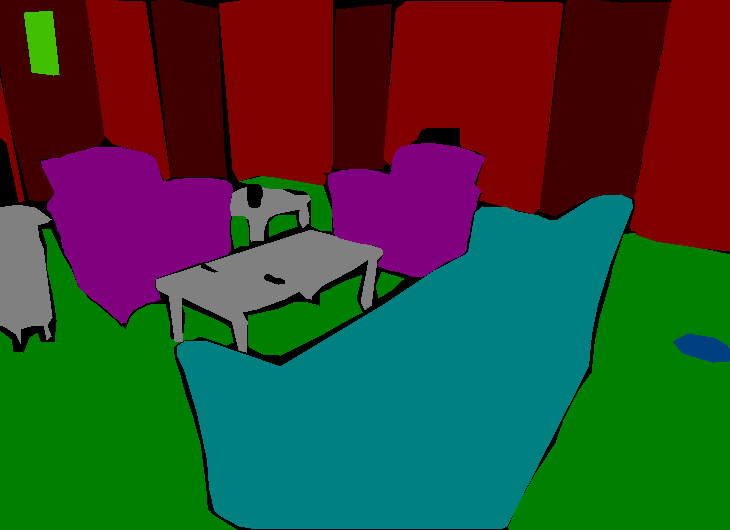}
        \includegraphics[width=0.11\textwidth]{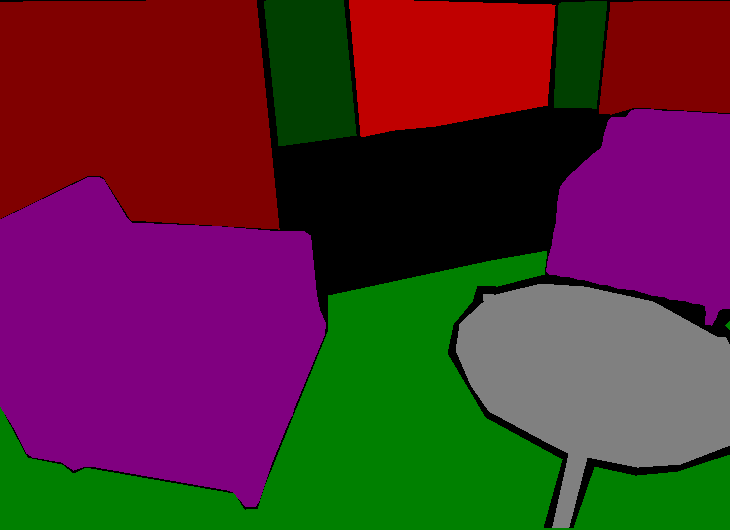} \\
        {\scriptsize Label Omission} & {\scriptsize Complete Labeling} \\[4pt]

        \includegraphics[width=0.11\textwidth]{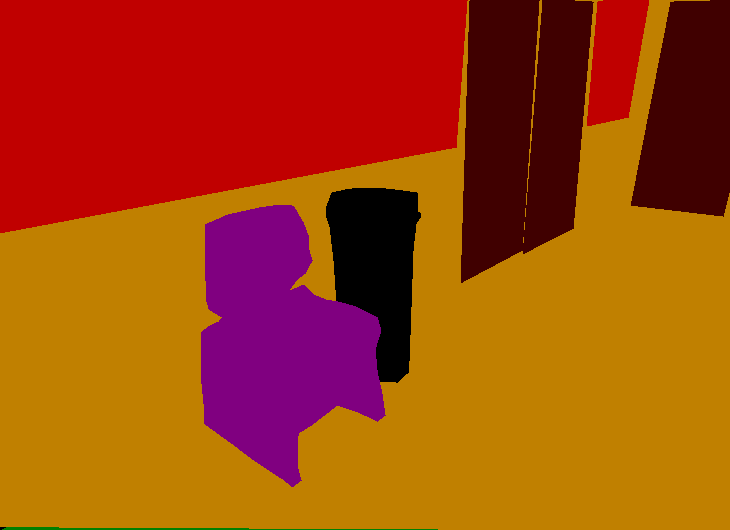}
        \includegraphics[width=0.11\textwidth]{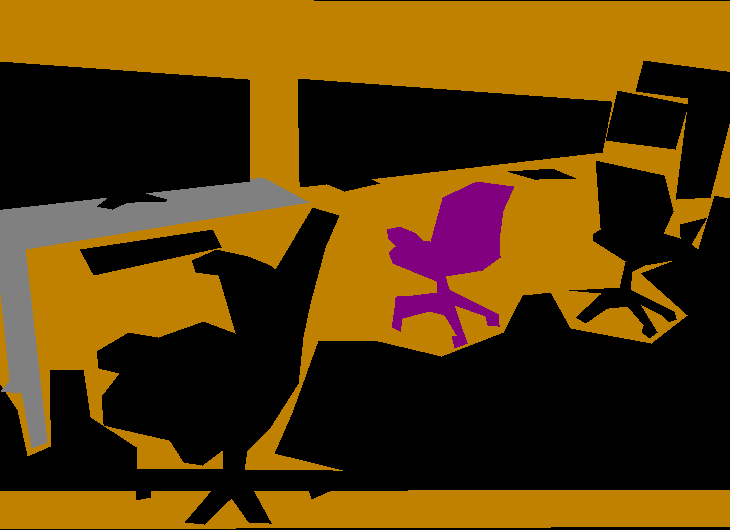} &
        \includegraphics[width=0.11\textwidth]{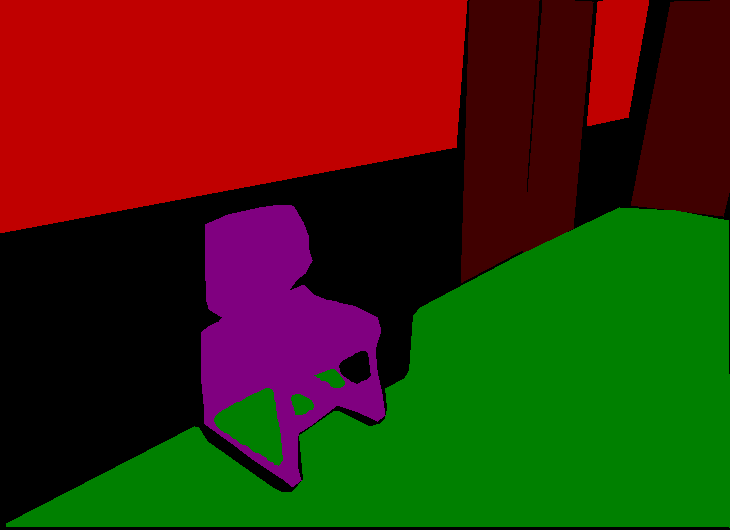}
        \includegraphics[width=0.11\textwidth]{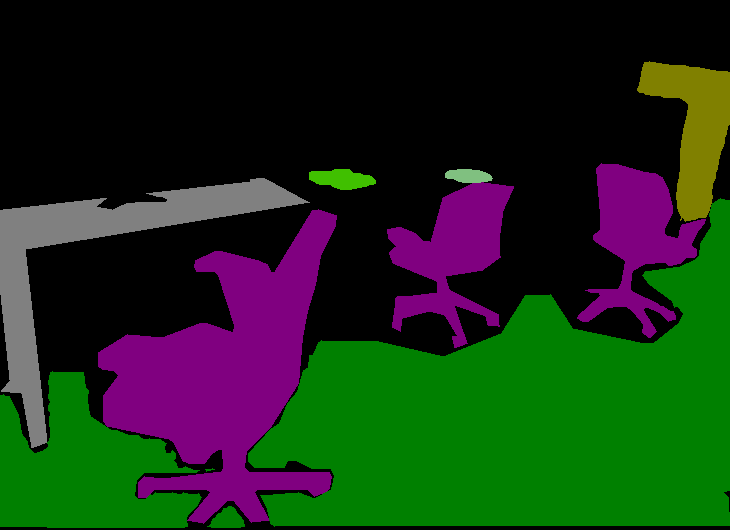} \\
        {\scriptsize Semantic Mislabeling} & {\scriptsize Accurate Categorization} \\[4pt]

        \includegraphics[width=0.11\textwidth]{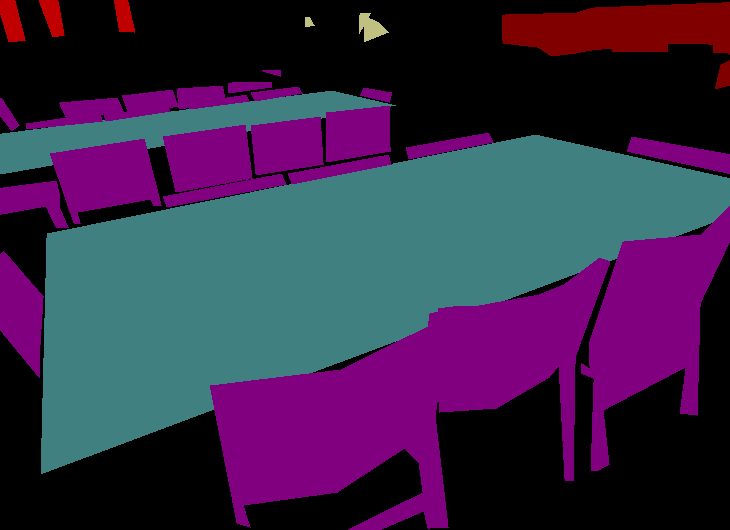}
        \includegraphics[width=0.11\textwidth]{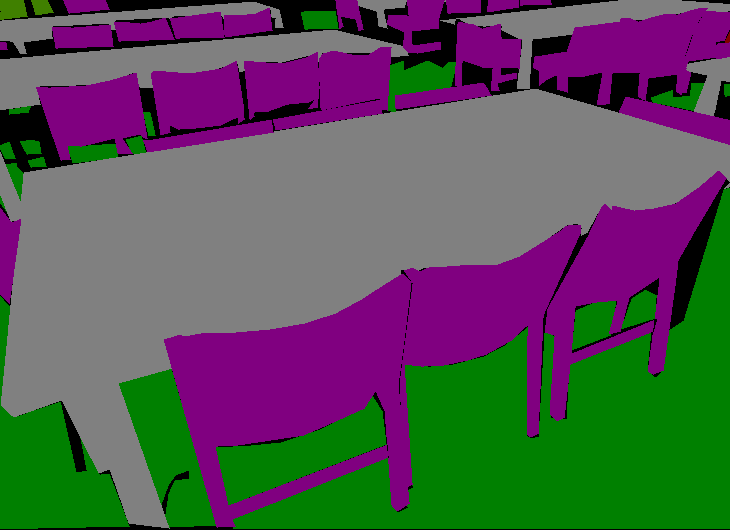} &
        \includegraphics[width=0.11\textwidth]{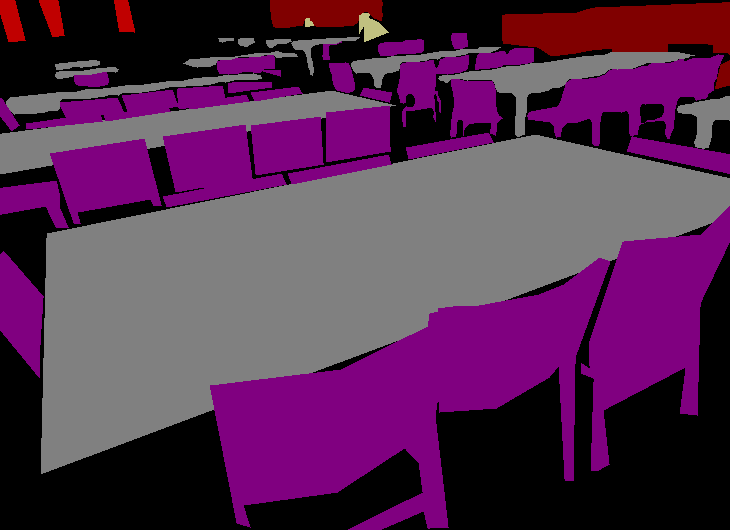}
        \includegraphics[width=0.11\textwidth]{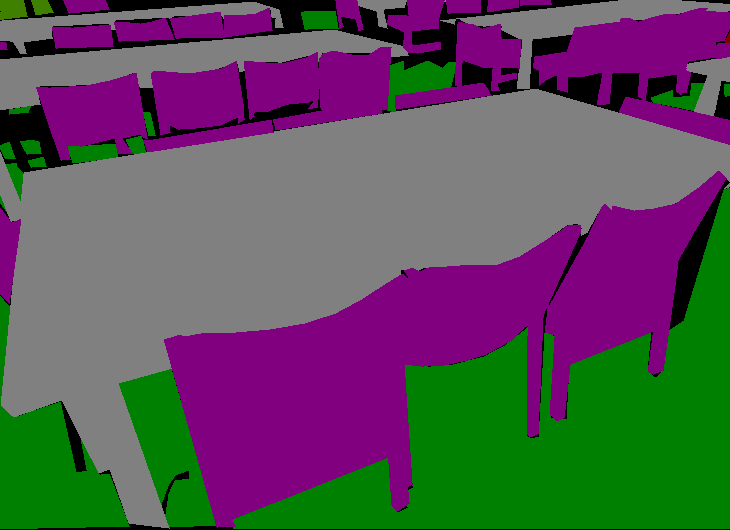} \\
        {\scriptsize Annotation Inconsistency} & {\scriptsize Labeling Consistency} \\[4pt]

        \includegraphics[width=0.11\textwidth]{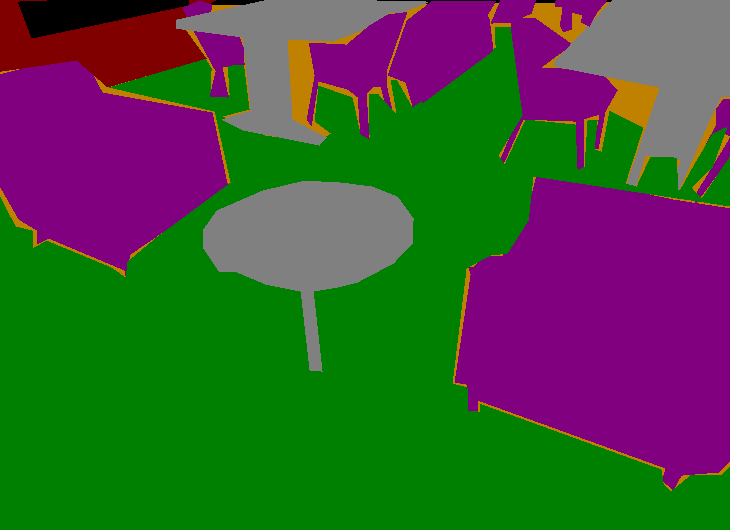}
        \includegraphics[width=0.11\textwidth]{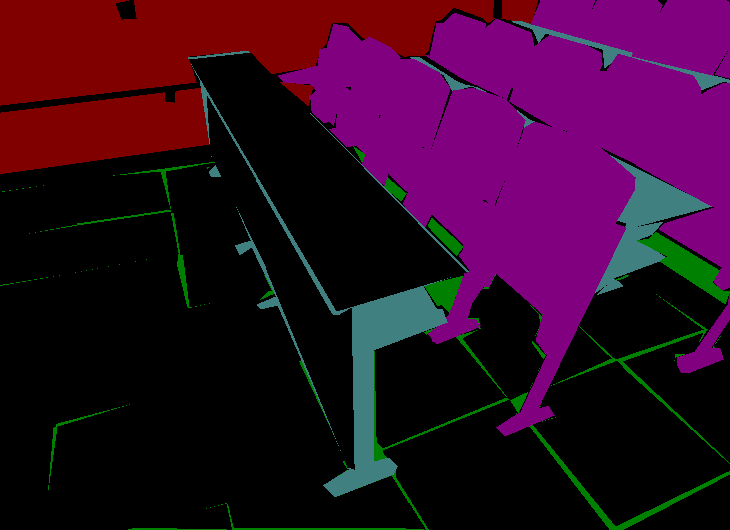} &
        \includegraphics[width=0.11\textwidth]{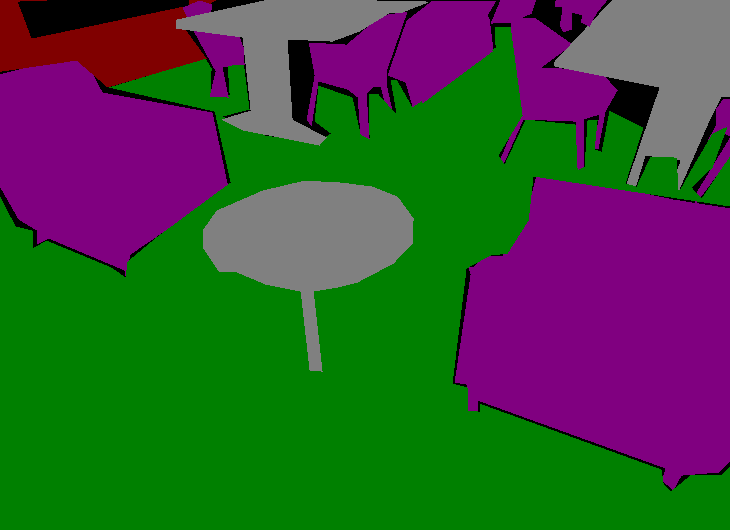}
        \includegraphics[width=0.11\textwidth]{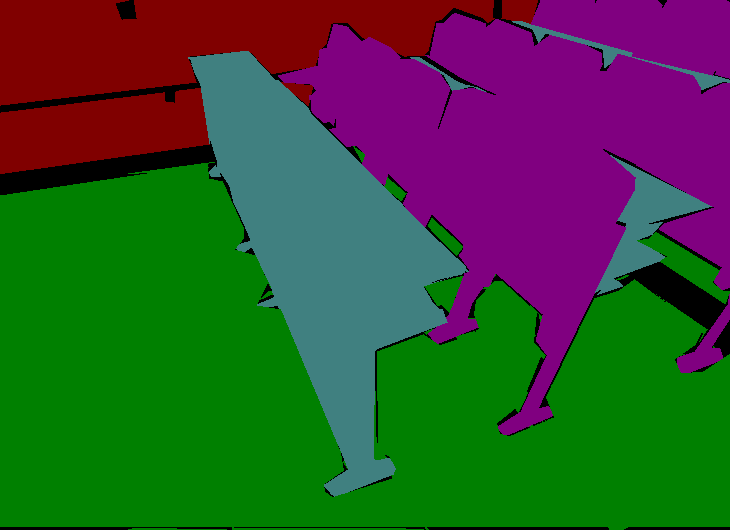} \\
        {\scriptsize Coarse Boundaries} & {\scriptsize Precise Boundaries} \\[8pt]

        {\scriptsize (a) Previous SUN RGB-D} & {\scriptsize (b) Our Re-annotation} \\
    \end{tabular}

    \caption{Comparison of annotation quality. (a) Previous dataset, (b) our high-quality results.}
    \label{anno_exa_4rows}
\end{figure}

\paragraph{Implementation Details} Following the standard training setting in GeminiFusion~\cite{jia2024geminifusion}, our backbone is SwinTransformer Large, we optimize our models using a weight decay of $0.01$. The entire training process spans $300$ epochs, utilizing a learning rate schedule divided into three equal stages of $100$ epochs each. Specifically, we set initial learning rates (LR) of $6 \times 10^{-5}$, $3 \times 10^{-5}$, and $1.5 \times 10^{-5}$ for the first, second, and final $100$ epochs, respectively. Throughout training, the learning rate in each stage is scheduled using the poly decay strategy with a power of $0.01$. Our models are trained on four NVIDIA H100 GPUs.

\subsection{Comparisons with the State of the Art}
\paragraph{Main Results} We compare our proposed SPF against a wide range of representative RGB-D semantic segmentation methods, including PGDENet~\cite{zhou2022pgdenet}, TokenFusion~\cite{wang2022multimodal}, GeminiFusion~\cite{jia2024geminifusion}, MultiMAE~\cite{bachmann2022multimae}, CMX~\cite{zhang2023cmx}, CMNeXt~\cite{zhang2023delivering}, DFormer v2~\cite{yin2025dformerv2} and DPLNet~\cite{dong2024efficient}, on the NYUv2~\cite{silberman2012indoor}, SUN RGB-D~\cite{song2015sun}, and our proposed RGBD20K datasets. For RGBD20K, we follow the original evaluation settings used by each method on SUN RGB-D for a fair comparison. The results are reported in Table~\ref{tab:comparison}.

\paragraph{SUN RGB-D Annotation Quality Analysis}
\label{sun rgbd analysis}
During dataset construction, we identified several annotation issues in SUN RGB-D, including category ambiguity, inaccurate labels, and noisy object boundaries. As shown in Figure~\ref{anno_exa_4rows}, we provide qualitative comparisons between the original and refined annotations to illustrate these issues. To further quantify their impact, we refine only the test set annotations and evaluate a pretrained DPLNet \cite{dong2024efficient} on the refined test set. This improves mIoU from 52.8 to \textbf{55.0}, demonstrating that annotation inconsistencies can substantially affect evaluation and lead to an underestimation of model performance. We replace the original annotations with our refined labels while retaining the original 37-category setting, yielding SUN RGB-D\textsuperscript{\dag}. We retrain representative RGB-D semantic segmentation models on this refined dataset. Under this setting, DPLNet \cite{dong2024efficient} achieves an mIoU of \textbf{59.0}, demonstrating that annotation refinement can substantially improve model performance when applied to both training and test data. As shown in Table~\ref{tab:comparison}, all evaluated methods consistently achieve higher performance on SUN RGB-D\textsuperscript{\dag}, demonstrating the effectiveness of our annotation refinement. These results highlight the significant impact of annotation quality on both evaluation and model training. Annotation inconsistencies in the training data can introduce noisy supervision, potentially impairing feature learning and model generalization. Based on these observations, we adopt the same annotation refinement principles throughout the construction of RGBD20K to ensure consistent and high-quality semantic annotations.

\begin{table*}[!t]
\centering
\small
\setlength{\tabcolsep}{6pt}
\renewcommand{\arraystretch}{1.0}
\caption{Comparison of RGBD20K to existing datasets using mean IoU.
\textsuperscript{\dag} denotes our re-annotation.}
\label{tab:comparison}
\begin{tabular}{lcccccc}
\toprule
& Backbone & Params & NYUv2 & SUN RGB-D & SUN RGB-D\textsuperscript{\dag} & RGBD20K \\
\midrule
PGDENet \cite{zhou2022pgdenet}
& ResNet-34 & 100.7M & 53.7 & 51.0 & 54.6 & 22.4 \\
TokenFusion \cite{wang2022multimodal}
& MiT-B3 & 45.9M & 54.2 & 51.4 & 56.2 & 32.8 \\
GeminiFusion \cite{jia2024geminifusion}
& Swin-L-384 & 369.2M & 60.2 & 54.6 & 61.5 & 45.4 \\
MultiMAE \cite{bachmann2022multimae}
& ViT-B & 95.2M & 56.0 & 51.1 & 55.7 & 42.2 \\
CMX \cite{zhang2023cmx}
& MiT-B5 & 181.1M & 56.9 & 52.4 & 55.7 & 43.9 \\
CMNeXt \cite{zhang2023delivering}
& MiT-B4 & 119.6M & 56.9 & 51.9 & 55.4 & 43.1 \\
DFormer v1 \cite{yin2023dformer}
& DFormer-L & 39.0M & 57.2 & 52.5 & 59.1 & 41.9 \\
DFormer v2 \cite{yin2025dformerv2}
& DFormer-v2-L & 95.5M & 58.4 & 53.3 & 59.7 & 44.1 \\
DPLNet \cite{dong2024efficient}
& MiT-B5 & 88.6M & 59.3 & 52.8 & 59.0 & 28.0 \\
SPF (Ours)
& Swin-L-384 & 416.4M & \textbf{60.5} & \textbf{55.0} & \textbf{62.2} & \textbf{46.3} \\
\bottomrule
\end{tabular}
\end{table*}

\subsection{Ablation Studies}
We conduct extensive ablation experiments on RGBD20K to validate the effectiveness of each component in our Score-Purified Fusion method.

\begin{table}[t]
    \centering
    \small
    \caption{Ablation study on different components.}
    \label{tab:ablation}
    \begin{tabular}{lc}
        \toprule
        \textbf{Structure} & \textbf{mIoU (\%)} \\ 
        \midrule
        Score-Purified Fusion & \textbf{46.3} \\ 
        \midrule
        without Projection & 45.7 {\color{dropcolor}(-0.6)} \\
        without Score & 45.5 {\color{dropcolor}(-0.8)} \\
        with Score (K) & 46.0 {\color{dropcolor}(-0.3)} \\
        with Score (V) & 45.9 {\color{dropcolor}(-0.4)} \\
        with Score (RGB) & 45.8 {\color{dropcolor}(-0.5)} \\
        with Score (Depth) & 46.0 {\color{dropcolor}(-0.3)} \\
        \bottomrule
    \end{tabular}
\end{table}

\textbf{Effect of RGB and Depth Alignment Projections.} We first analyze the impact of cross-modal alignment. We remove the specific projection layers for the RGB and depth modalities, processing both branches independently without explicit alignment. As shown in Table~\ref{tab:ablation}, this variant leads to a noticeable performance drop, demonstrating that cross-modal projection is essential for effective feature alignment and interaction.

\textbf{Evaluation of the Score-Guided Enhancement Strategy.} We conduct a comprehensive analysis of our method by evaluating the application of our score to enhance the Key ($K$) and Value ($V$) representations across different modalities and components. First, restricting the score enhancement to either the RGB branch or the depth branch alone, while reverting the other to a simple summation, yields inferior results, confirming that bidirectional enhancement is essential for balanced cross-modal learning. Second, independently applying the purification score to either the Key or Value alone while using simple summation for the other underperforms the complete model. The best results are achieved by jointly enhancing both components, demonstrating that consistent refinement of both attention computation (Key) and feature retrieval (Value) is critical. Results are shown in Table~\ref{tab:ablation}.

\textbf{Effect of RGB and Depth noise selection.} We also conduct ablation studies on noise selection strategies, with results reported in Table \ref{tab:ablation_noise}. Our findings show that the best performance is achieved by introducing a learnable parameter into the key, where this parameter is independently defined for each layer. Replacing this design with a simple \textit{Multiply} operation leads to a slight performance drop of 0.2\% (46.1\% mIoU). This observation is consistent with the findings of GeminiFusion~\cite{jia2024geminifusion}, further demonstrating the effectiveness of layer-specific learnable noise for cross-modal alignment.

\textbf{Effect of Relation Arbiter Design.} We investigate the architectural design of the Relation Arbiter on the RGBD20K dataset, with results summarized in Table \ref{tab:ablation_arbiter}. Our experiments compare different transformation layers and activation functions to determine the most effective way to modulate cross-modal relationships. We find that a 2-layer MLP combined with a Sigmoid activation achieves the highest performance, reaching 46.3\% mIoU.

\begin{table}[!t] 
    \centering
    \small
    
    \begin{minipage}{0.48\textwidth}
        \centering
        \caption{Ablation about the noise selection on the RGBD20K dataset.}
        \label{tab:ablation_noise}
        \begin{tabular}{lc}
            \toprule
            \textbf{Structure} & \textbf{mIoU (\%)} \\ \midrule
            Learnable Noise, Add  & \textbf{46.3} \\ \midrule
            Learnable Noise, Multiply & 46.1 {\color{dropcolor}(-0.2)} \\
            Random Gaussian Noise, Add  & 45.3 {\color{dropcolor}(-1.0)} \\
            Random Gaussian Noise, Multiply & 45.5 {\color{dropcolor}(-0.8)} \\
            \bottomrule
        \end{tabular}
    \end{minipage}
    \hfill 
    \begin{minipage}{0.48\textwidth}
        \centering
        \caption{Ablation about the Relation Arbiter on the RGBD20K dataset.}
        \label{tab:ablation_arbiter}
        \begin{tabular}{lc}
            \toprule
            \textbf{Structure} & \textbf{mIoU (\%)} \\ \midrule
            2layer-MLP + Sigmoid  & \textbf{46.3} \\ \midrule
            2layer-MLP + Softmax & 45.8 {\color{dropcolor}(-0.5)} \\
            1 $\times$ 1 CNN + Sigmoid & 45.5 {\color{dropcolor}(-0.8)} \\
            3 $\times$ 3 CNN + Sigmoid & 45.4 {\color{dropcolor}(-0.9)} \\
            \bottomrule
        \end{tabular}
    \end{minipage}
\end{table}

\section{CONCLUSIONS}

We introduce \textbf{RGBD20K}, a large-scale benchmark for RGB-D semantic segmentation. To bridge the gap in taxonomic diversity and annotation quality, RGBD20K provides 20,000 RGB-D image pairs annotated across 160 fine-grained categories. As one of the most comprehensive RGB-D benchmarks to date, it establishes a high-fidelity foundation for training general-purpose perception models. Furthermore, its dense, depth-aligned ground truth enables a deeper exploration of multimodal synergy, addressing the limitations of RGB-only approaches in complex scenes. To set a robust baseline for future research, we extensively evaluate representative segmentation models on RGBD20K. Additionally, we propose the score-purified fusion method, which achieves state-of-the-art performance across all evaluated benchmarks, demonstrating its effectiveness. By releasing RGBD20K, we aim to advance next-generation RGB-D semantic perception for robotic and autonomous systems.

\bibliographystyle{IEEEtran}
\bibliography{IEEEabrv, main}

@inproceedings{silberman2012indoor,
  title={Indoor segmentation and support inference from rgbd images},
  author={Silberman, Nathan and Hoiem, Derek and Kohli, Pushmeet and Fergus, Rob},
  booktitle={European conference on computer vision},
  pages={746--760},
  year={2012},
  organization={Springer}
}

@inproceedings{song2015sun,
  title={Sun rgb-d: A rgb-d scene understanding benchmark suite},
  author={Song, Shuran and Lichtenberg, Samuel P and Xiao, Jianxiong},
  booktitle={Proceedings of the IEEE conference on computer vision and pattern recognition},
  pages={567--576},
  year={2015}
}

@inproceedings{dai2017scannet,
  title={Scannet: Richly-annotated 3d reconstructions of indoor scenes},
  author={Dai, Angela and Chang, Angel X and Savva, Manolis and Halber, Maciej and Funkhouser, Thomas and Nie{\ss}ner, Matthias},
  booktitle={Proceedings of the IEEE conference on computer vision and pattern recognition},
  pages={5828--5839},
  year={2017}
}

@article{chang2017matterport3d,
  title={Matterport3d: Learning from rgb-d data in indoor environments},
  author={Chang, Angel and Dai, Angela and Funkhouser, Thomas and Halber, Maciej and Niessner, Matthias and Savva, Manolis and Song, Shuran and Zeng, Andy and Zhang, Yinda},
  journal={arXiv preprint arXiv:1709.06158},
  year={2017}
}

@article{armeni2017joint,
  title={Joint 2d-3d-semantic data for indoor scene understanding},
  author={Armeni, Iro and Sax, Sasha and Zamir, Amir R and Savarese, Silvio},
  journal={arXiv preprint arXiv:1702.01105},
  year={2017}
}

@misc{Li2025realsee3d_data,
  doi = {10.5281/zenodo.17826243},
  url = {https://doi.org/10.5281/zenodo.17826243},
  author = {Li, Linyuan and Wu, Yan and Li, Xi and Wang, Lingli and Rao, Tong and Zhou, Jie and Pan, Cihui and Hui, Xinchen},
  title = {Realsee3D: A Large-Scale Multi-View RGB-D Dataset of Indoor Scenes (Version 1.0)},
  publisher = {Zenodo},
  year = {2025}
}

@article{zhou2022pgdenet,
  title={PGDENet: Progressive guided fusion and depth enhancement network for RGB-D indoor scene parsing},
  author={Zhou, Wujie and Yang, Enquan and Lei, Jingsheng and Wan, Jian and Yu, Lu},
  journal={IEEE Transactions on Multimedia},
  volume={25},
  pages={3483--3494},
  year={2022},
  publisher={IEEE}
}

@inproceedings{wang2022multimodal,
  title={Multimodal token fusion for vision transformers},
  author={Wang, Yikai and Chen, Xinghao and Cao, Lele and Huang, Wenbing and Sun, Fuchun and Wang, Yunhe},
  booktitle={Proceedings of the IEEE/CVF conference on computer vision and pattern recognition},
  pages={12186--12195},
  year={2022}
}

@article{jia2024geminifusion,
  title={Geminifusion: Efficient pixel-wise multimodal fusion for vision transformer},
  author={Jia, Ding and Guo, Jianyuan and Han, Kai and Wu, Han and Zhang, Chao and Xu, Chang and Chen, Xinghao},
  journal={arXiv preprint arXiv:2406.01210},
  year={2024}
}

@inproceedings{bachmann2022multimae,
  title={Multimae: Multi-modal multi-task masked autoencoders},
  author={Bachmann, Roman and Mizrahi, David and Atanov, Andrei and Zamir, Amir},
  booktitle={European conference on computer vision},
  pages={348--367},
  year={2022},
  organization={Springer}
}

@article{yin2023dformer,
  title={Dformer: Rethinking rgbd representation learning for semantic segmentation},
  author={Yin, Bowen and Zhang, Xuying and Li, Zhongyu and Liu, Li and Cheng, Ming-Ming and Hou, Qibin},
  journal={arXiv preprint arXiv:2309.09668},
  year={2023}
}

@inproceedings{yin2025dformerv2,
  title={Dformerv2: Geometry self-attention for rgbd semantic segmentation},
  author={Yin, Bo-Wen and Cao, Jiao-Long and Cheng, Ming-Ming and Hou, Qibin},
  booktitle={Proceedings of the IEEE/CVF Conference on Computer Vision and Pattern Recognition},
  pages={19345--19355},
  year={2025}
}

@inproceedings{dong2024efficient,
  title={Efficient multimodal semantic segmentation via dual-prompt learning},
  author={Dong, Shaohua and Feng, Yunhe and Yang, Qing and Huang, Yan and Liu, Dongfang and Fan, Heng},
  booktitle={2024 IEEE/RSJ International Conference on Intelligent Robots and Systems (IROS)},
  pages={14196--14203},
  year={2024},
  organization={IEEE}
}

@article{zhang2023cmx,
  title={CMX: Cross-modal fusion for RGB-X semantic segmentation with transformers},
  author={Zhang, Jiaming and Liu, Huayao and Yang, Kailun and Hu, Xinxin and Liu, Ruiping and Stiefelhagen, Rainer},
  journal={IEEE Transactions on intelligent transportation systems},
  volume={24},
  number={12},
  pages={14679--14694},
  year={2023},
  publisher={IEEE}
}

@inproceedings{zhang2023delivering,
  title={Delivering arbitrary-modal semantic segmentation},
  author={Zhang, Jiaming and Liu, Ruiping and Shi, Hao and Yang, Kailun and Rei{\ss}, Simon and Peng, Kunyu and Fu, Haodong and Wang, Kaiwei and Stiefelhagen, Rainer},
  booktitle={Proceedings of the IEEE/CVF Conference on Computer Vision and Pattern Recognition},
  pages={1136--1147},
  year={2023}
}

@inproceedings{cai2025keep,
  title={Keep the Balance: A Parameter-Efficient Symmetrical Framework for RGB+ X Semantic Segmentation},
  author={Cai, Jiaxin and Su, Jingze and Li, Qi and Yang, Wenjie and Wang, Shu and Zhao, Tiesong and He, Shengfeng and Liu, Wenxi},
  booktitle={Proceedings of the Computer Vision and Pattern Recognition Conference},
  pages={10587--10598},
  year={2025}
}

@inproceedings{ha2017mfnet,
  title={MFNet: Towards real-time semantic segmentation for autonomous vehicles with multi-spectral scenes},
  author={Ha, Qishen and Watanabe, Kohei and Karasawa, Takumi and Ushiku, Yoshitaka and Harada, Tatsuya},
  booktitle={2017 IEEE/RSJ International Conference on Intelligent Robots and Systems (IROS)},
  pages={5108--5115},
  year={2017},
  organization={IEEE}
}

@inproceedings{shivakumar2020pst900,
  title={Pst900: Rgb-thermal calibration, dataset and segmentation network},
  author={Shivakumar, Shreyas S and Rodrigues, Neil and Zhou, Alex and Miller, Ian D and Kumar, Vijay and Taylor, Camillo J},
  booktitle={2020 IEEE international conference on robotics and automation (ICRA)},
  pages={9441--9447},
  year={2020},
  organization={IEEE}
}

@inproceedings{ji2023semanticrt,
  title={Semanticrt: A large-scale dataset and method for robust semantic segmentation in multispectral images},
  author={Ji, Wei and Li, Jingjing and Bian, Cheng and Zhang, Zhicheng and Cheng, Li},
  booktitle={Proceedings of the 31st ACM International Conference on Multimedia},
  pages={3307--3316},
  year={2023}
}

@inproceedings{ji2023multispectral,
  title={Multispectral video semantic segmentation: A benchmark dataset and baseline},
  author={Ji, Wei and Li, Jingjing and Bian, Cheng and Zhou, Zongwei and Zhao, Jiaying and Yuille, Alan L and Cheng, Li},
  booktitle={Proceedings of the IEEE/CVF Conference on Computer Vision and Pattern Recognition},
  pages={1094--1104},
  year={2023}
}

@article{zhou2022mtanet,
  title={MTANet: Multitask-aware network with hierarchical multimodal fusion for RGB-T urban scene understanding},
  author={Zhou, Wujie and Dong, Shaohua and Lei, Jingsheng and Yu, Lu},
  journal={IEEE Transactions on Intelligent Vehicles},
  volume={8},
  number={1},
  pages={48--58},
  year={2022},
  publisher={IEEE}
}

@inproceedings{zhou2022edge,
  title={Edge-aware guidance fusion network for rgb--thermal scene parsing},
  author={Zhou, Wujie and Dong, Shaohua and Xu, Caie and Qian, Yaguan},
  booktitle={Proceedings of the AAAI conference on artificial intelligence},
  volume={36},
  pages={3571--3579},
  year={2022}
}

@article{dong2023egfnet,
  title={EGFNet: Edge-aware guidance fusion network for RGB--thermal urban scene parsing},
  author={Dong, Shaohua and Zhou, Wujie and Xu, Caie and Yan, Weiqing},
  journal={IEEE Transactions on Intelligent Transportation Systems},
  volume={25},
  number={1},
  pages={657--669},
  year={2023},
  publisher={IEEE}
}

@article{zhou2023cacfnet,
  title={CACFNet: Cross-modal attention cascaded fusion network for RGB-T urban scene parsing},
  author={Zhou, Wujie and Dong, Shaohua and Fang, Meixin and Yu, Lu},
  journal={IEEE Transactions on Intelligent Vehicles},
  volume={9},
  number={1},
  pages={1919--1929},
  year={2023},
  publisher={IEEE}
}

@article{dong2022gebnet,
  title={GEBNet: Graph-enhancement branch network for RGB-T scene parsing},
  author={Dong, Shaohua and Zhou, Wujie and Qian, Xiaohong and Yu, Lu},
  journal={IEEE Signal Processing Letters},
  volume={29},
  pages={2273--2277},
  year={2022},
  publisher={IEEE}
}

@inproceedings{mei2021depth,
  title={Depth-aware mirror segmentation},
  author={Mei, Haiyang and Dong, Bo and Dong, Wen and Peers, Pieter and Yang, Xin and Zhang, Qiang and Wei, Xiaopeng},
  booktitle={Proceedings of the IEEE/CVF conference on computer vision and pattern recognition},
  pages={3044--3053},
  year={2021}
}

@article{lin2024vidsod,
  title={Vidsod-100: A new dataset and a baseline model for rgb-d video salient object detection},
  author={Lin, Junhao and Zhu, Lei and Shen, Jiaxing and Fu, Huazhu and Zhang, Qing and Wang, Liansheng},
  journal={International Journal of Computer Vision},
  volume={132},
  number={11},
  pages={5173--5191},
  year={2024},
  publisher={Springer}
}

@inproceedings{yan2021depthtrack,
  title={Depthtrack: Unveiling the power of rgbd tracking},
  author={Yan, Song and Yang, Jinyu and K{\"a}pyl{\"a}, Jani and Zheng, Feng and Leonardis, Ale{\v{s}} and K{\"a}m{\"a}r{\"a}inen, Joni-Kristian},
  booktitle={Proceedings of the IEEE/CVF international conference on computer vision},
  pages={10725--10733},
  year={2021}
}

@inproceedings{zhu2023rgbd1k,
  title={RGBD1K: A large-scale dataset and benchmark for RGB-D object tracking},
  author={Zhu, Xue-Feng and Xu, Tianyang and Tang, Zhangyong and Wu, Zucheng and Liu, Haodong and Yang, Xiao and Wu, Xiao-Jun and Kittler, Josef},
  booktitle={Proceedings of the AAAI Conference on Artificial Intelligence},
  volume={37},
  pages={3870--3878},
  year={2023}
}

@inproceedings{zhao2023arkittrack,
  title={Arkittrack: a new diverse dataset for tracking using mobile RGB-D data},
  author={Zhao, Haojie and Chen, Junsong and Wang, Lijun and Lu, Huchuan},
  booktitle={Proceedings of the IEEE/CVF Conference on Computer Vision and Pattern Recognition},
  pages={5126--5135},
  year={2023}
}

@article{cho2021diml,
  title={Diml/cvl rgb-d dataset: 2m rgb-d images of natural indoor and outdoor scenes},
  author={Cho, Jaehoon and Min, Dongbo and Kim, Youngjung and Sohn, Kwanghoon},
  journal={arXiv preprint arXiv:2110.11590},
  year={2021}
}

@article{xie2021segformer,
  title={SegFormer: Simple and efficient design for semantic segmentation with transformers},
  author={Xie, Enze and Wang, Wenhai and Yu, Zhiding and Anandkumar, Anima and Alvarez, Jose M and Luo, Ping},
  journal={Advances in neural information processing systems},
  volume={34},
  pages={12077--12090},
  year={2021}
}

@inproceedings{liu2021swin,
  title={Swin transformer: Hierarchical vision transformer using shifted windows},
  author={Liu, Ze and Lin, Yutong and Cao, Yue and Hu, Han and Wei, Yixuan and Zhang, Zheng and Lin, Stephen and Guo, Baining},
  booktitle={Proceedings of the IEEE/CVF international conference on computer vision},
  pages={10012--10022},
  year={2021}
}

@article{dosovitskiy2020image,
  title={An image is worth 16x16 words: Transformers for image recognition at scale},
  author={Dosovitskiy, Alexey and Beyer, Lucas and Kolesnikov, Alexander and Weissenborn, Dirk and Zhai, Xiaohua and Unterthiner, Thomas and Dehghani, Mostafa and Minderer, Matthias and Heigold, Georg and Gelly, Sylvain and others},
  journal={arXiv preprint arXiv:2010.11929},
  year={2020}
}

@article{zhou2019semantic,
  title={Semantic understanding of scenes through the ade20k dataset},
  author={Zhou, Bolei and Zhao, Hang and Puig, Xavier and Xiao, Tete and Fidler, Sanja and Barriuso, Adela and Torralba, Antonio},
  journal={International journal of computer vision},
  volume={127},
  number={3},
  pages={302--321},
  year={2019},
  publisher={Springer}
}

@article{everingham2010pascal,
  title={The pascal visual object classes (voc) challenge},
  author={Everingham, Mark and Van Gool, Luc and Williams, Christopher KI and Winn, John and Zisserman, Andrew},
  journal={International journal of computer vision},
  volume={88},
  number={2},
  pages={303--338},
  year={2010},
  publisher={Springer}
}

@inproceedings{silberman2011indoor,
  title={Indoor scene segmentation using a structured light sensor},
  author={Silberman, Nathan and Fergus, Rob},
  booktitle={2011 IEEE international conference on computer vision workshops (ICCV workshops)},
  pages={601--608},
  year={2011},
  organization={IEEE}
}

@article{li2025dmtrack,
  title={DMTrack: Spatio-Temporal Multimodal Tracking via Dual-Adapter},
  author={Li, Weihong and Dong, Shaohua and Lu, Haonan and Zhang, Yanhao and Fan, Heng and Zhang, Libo},
  journal={arXiv preprint arXiv:2508.01592},
  year={2025}
}

@article{peng2024vasttrack,
  title={Vasttrack: Vast category visual object tracking},
  author={Peng, Liang and Gao, Junyuan and Liu, Xinran and Li, Weihong and Dong, Shaohua and Zhang, Zhipeng and Fan, Heng and Zhang, Libo},
  journal={Advances in Neural Information Processing Systems},
  volume={37},
  pages={130797--130818},
  year={2024}
}

@article{dong2024loretrack,
  title={Loretrack: efficient and accurate low-resolution transformer tracking},
  author={Dong, Shaohua and Feng, Yunhe and Yang, Qing and Lin, Yuewei and Fan, Heng},
  journal={arXiv preprint arXiv:2405.17660},
  year={2024}
}

\end{document}